\pdfoutput=1
\documentclass[10pt,twocolumn,letterpaper]{article}

\usepackage{cvpr}              

\usepackage{graphicx}
\usepackage{amsmath}
\usepackage{amssymb}
\usepackage{booktabs}
\usepackage{epsfig}
\usepackage{multirow}
\usepackage{enumitem}
\usepackage{caption}
\usepackage{subcaption}
\usepackage{xcolor}
\usepackage{amsthm}
\usepackage[ruled,vlined]{algorithm2e}
\usepackage{tabularx,booktabs}
\usepackage{hhline}
\usepackage{pifont}
\usepackage{diagbox}
\usepackage{makecell}
\usepackage[accsupp]{axessibility}

\usepackage[pagebackref,breaklinks,colorlinks]{hyperref}

\usepackage[capitalize]{cleveref}
\crefname{section}{Sec.}{Secs.}
\Crefname{section}{Section}{Sections}
\Crefname{table}{Table}{Tables}
\crefname{table}{Tab.}{Tabs.}

\setlist[itemize]{noitemsep, nolistsep}

\newcommand{\Our}{GGN}
\newcommand{\Ours}{GGNs}

\def\cvprPaperID{9506} 
\def\confName{CVPR}
\def\confYear{2022}

\begin{document}


\title{Open-World Instance Segmentation:\\ Exploiting Pseudo Ground Truth From Learned Pairwise Affinity}


\author{Weiyao Wang\textsuperscript{\textdagger}, Matt Feiszli\textsuperscript{\textdagger}, Heng Wang\textsuperscript{\textdagger}, Jitendra Malik\textsuperscript{\textdagger \S}, Du Tran\textsuperscript{\textdagger}\\
\textsuperscript{\textdagger} Meta AI Research \ \  \textsuperscript{\S} UC Berkeley\\
{\tt\small \{weiyaowang,mdf,hengwang,trandu\}@fb.com, malik@berkeley.edu} \\
{\href{https://sites.google.com/view/generic-grouping/}{sites.google.com/view/generic-grouping/}}
}
\twocolumn[{%
\renewcommand\twocolumn[1][]{#1}%
\maketitle
\begin{center}
    \centering
    \captionsetup{type=figure}
    \includegraphics[width=\textwidth]{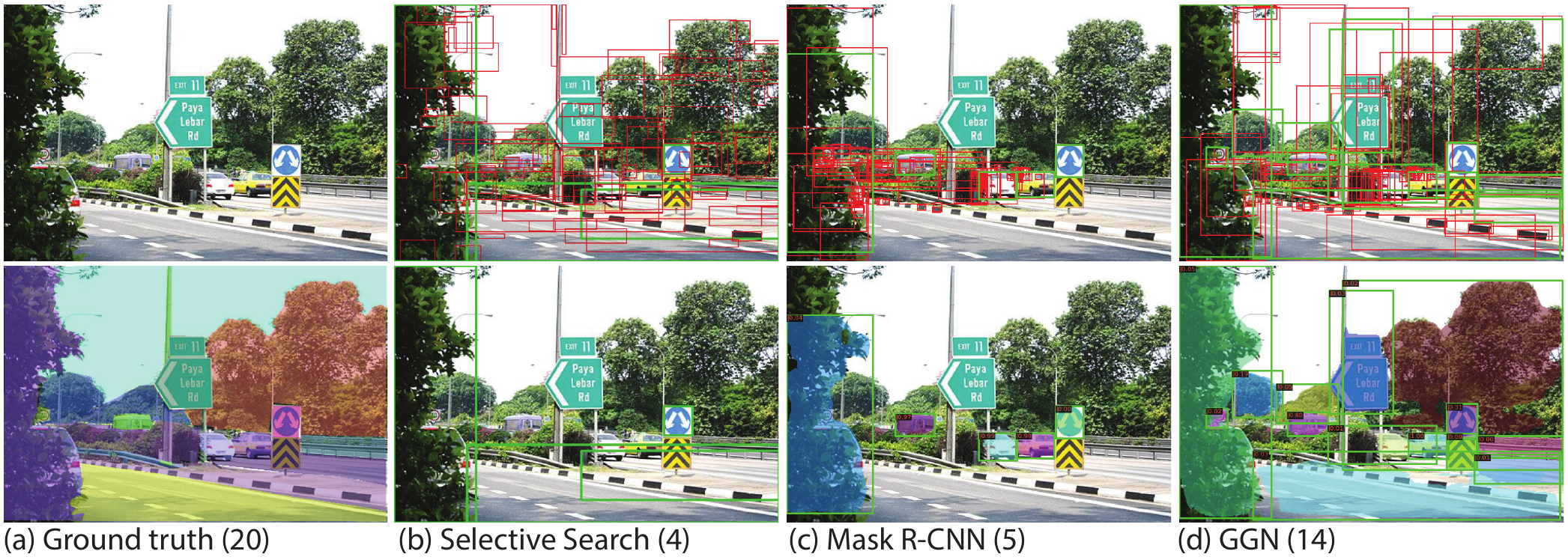}
    \vspace{-8pt}
    \caption{{\bf Comparison of GGN with different baselines}. (a) an input image from ADE20K~\cite{8100027} (upper row) with the ground truth object masks overlaid (lower row). Three different approaches: (b) Selective Search (SS)~\cite{SelectiveSearch}, (c) Mask R-CNN~\cite{8237584}, and (d) our Generic Grouping Network (\Our{}) are applied on the same image to predict the top $100$ proposals. The upper images provide all top-100 proposals predicted by three approaches with false and true positive proposals visualized in red and green boxes, respectively. The lower images provide only true positive proposals. The number of true positive proposals or ground truth objects are denoted in parentheses. Among 20 ground truth objects, SS recalls only 4, Mask R-CNN detects 5, and \Our{} retrieves 14. SS is a bottom-up non-parametric approach, thus has no notion of objectness. Mask R-CNN can make whole object proposals; however it still fails to detect objects that are not seen during training. Our \Our{} can predict whole object proposals and generalize to unseen categories.}
    \label{fig:intro}
\end{center}%
}]

\begin{abstract}

Open-world instance segmentation is the task of grouping pixels into object instances without any pre-determined taxonomy. This is challenging, as state-of-the-art methods rely on explicit class semantics obtained from large labeled datasets, and out-of-domain evaluation performance drops significantly. Here we propose a novel approach for mask proposals, Generic Grouping Networks (\Ours{}), constructed without semantic supervision. Our approach combines a local measure of pixel affinity with instance-level mask supervision, producing a training regimen designed to make the model as generic as the data diversity allows.  We introduce a method for predicting Pairwise Affinities (PA), a learned local relationship between pairs of pixels. PA generalizes very well to unseen categories. From PA we construct a large set of pseudo-ground-truth instance masks; combined with human-annotated instance masks we train \Ours{} and significantly outperform the SOTA on open-world instance segmentation on various benchmarks including COCO, LVIS, ADE20K, and UVO. Code is available on project \href{https://sites.google.com/view/generic-grouping/}{website}.



\end{abstract}

\section{Introduction}
\label{sec:intro}

Instance segmentation is the task of grouping pixels into object instances~\cite{8237584}. In the closed-world setup, the task is to detect and segment objects from a predefined taxonomy. In contrast, the open-world setting requires segmenting objects of arbitrary categories. For a model trained in a closed-world setup, this means segmenting not only the ``seen'' categories (those presented at training time) but also the ``unseen'' categories (not seen during training)~\cite{uvo,oln}.

There is generally a large performance gap between the seen and unseen domains. Leading computer vision systems today have tightly coupled recognition and segmentation; these systems are unable to segment out objects that they cannot recognize (e.g. Fig~\ref{fig:intro} (c)). Comparing Average Recall (AR@100) of Mask R-CNN~\cite{8237584} trained on 80 COCO~\cite{Lin2014MicrosoftCC} classes vs a subset of 20 classes, AR@100 of 60 classes out of training taxonomy drops from $49.6\%$ to $19.9\%$ when no mask of these classes is provided in training data. The ``unseen'' gap remains large if we train on larger taxonomy (e.g., 1,000+ classes in training data) Table~\ref{tab:diff_setup}). In contrast, humans can readily group and segment objects which they cannot categorize - few of us can identify the 6500 Passerine bird species, but we can readily segment out a perching bird from a tree branch. Or use another often-quoted example: our familiarity with a generic quadruped body plan enables us to segment out horses, donkeys and zebras, and even an okapi when first encountered. 

On the other hand, models which were common in computer vision in 2000-2015 (e.g., ~\cite{SelectiveSearch,Achanta10slicsuperpixels,GBH,SEEDS3D,StreamGBH,ArbelaezPBMM14}), before deep learning for supervised object detection took off, were quite category-agnostic. They didn't work as well as, say, Mask R-CNN on a category for which it has trained, but they worked across the board (e.g. Figure~\ref{fig:intro} (b)). The goal was to come up with a moderately sized set of object proposals which included the true objects. The emphasis was on recall; precision was secondary. MCG~\cite{ArbelaezPBMM14} is an illustrative example. It starts with local grouping which produces a set of elementary regions of coherent color and texture, ``super-pixels''. These typically over-segment objects; e.g. a person might be broken up into a face, a torso, legs, parts of clothing, shadows, etc. MCG then assembles regions into objects by considering various groupings of regions, and ranks them on some ``objectness'' score. While some learning is involved in both edge detection and objectness ranking, the method works primarily with hand-crafted features and a small number of parameters, quite unlike the deep learning zeitgeist.

How do we get the best of both worlds? A modern instance segmentation system (eg. Mask R-CNN) would do well if given comprehensive training data containing a large number of examples from all visual categories. While we have a practically infinite supply of raw natural images, obtaining mask annotation is very expensive. Multiple approaches have emerged to handle this data problem. Self-supervised learning~\cite{Pathak2016ContextEF,chen2020simple,He2020MomentumCF,chen2020mocov2} is the most well-known; self-learning~\cite{1053799,10.3115/981658.981684,10.5555/1864519.1864542,Zoph2020RethinkingST} is another approach, based on the classical idea of adding high-confidence guessed labels to previously unlabeled data, and then combining this ``pseudo ground truth'' data with real ground truth. We exploit this second strategy.

\begin{figure*}
  \centering
    \includegraphics[width=0.95\linewidth]{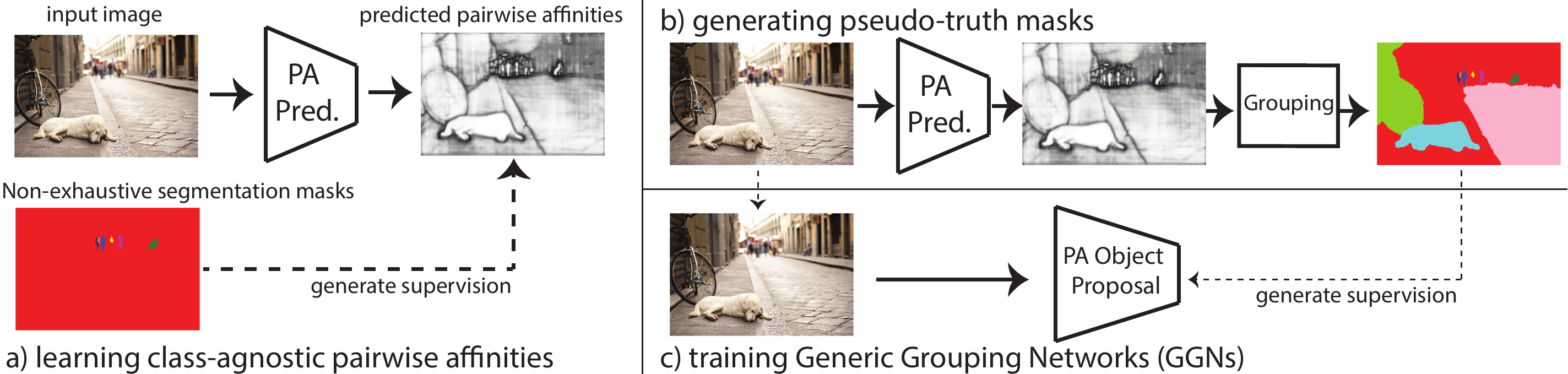}
  \vspace{-8pt}
  \caption{{\bf Overview of our approach}. (a) First, a Pairwise-Affinity Predictor (\emph{PA Pred.}) is trained to predict pairwise affinities using non-exhaustive segmentation masks as supervision. (b) Once trained, \emph{PA Pred.} is used to predict pairwise affinities of the images. A grouping module is then applied on the predicted pairwise affinity maps to generate pseudo-ground-truth masks. (c) A class-agnostic generic object proposal network (e.g., class-agnostic Mask R-CNN) is trained end-to-end using a combination of GT and generated pseudo-GT masks.}
  \label{fig:overview}
  \vspace{-5pt}
\end{figure*}

Our approach begins with a learned pairwise affinity predictor (Figure~\ref{fig:overview}a), followed by a module which extracts and ranks segments (Figure~\ref{fig:overview}b, essentially a very simplified version of MCG~\cite{ArbelaezPBMM14}).  We can run this on any image dataset without using annotations; we extract the highest ranked segments as ``pseudo ground truth'' candidate objects. This is a large and category-agnostic set; we add it to our (much smaller) datasets of curated annotations, to train a Mask R-CNN instance segmentation module.  Ideally this model should become more generic and class-agnostic (Figure~\ref{fig:overview}c). Indeed, this simple approach produces impressive gains compared to closed-world training on the same backbone (Mask R-CNN) (Table~\ref{tab:diff_setup}, Table~\ref{tab:compare_oln}, Table~\ref{tab:compare_wild}): \textbf{+11\%} on VOC to Non-VOC cross-category evaluation, \textbf{+3.9\%} on COCO to LVIS cross-category evaluation, \textbf{+5.8\%} on COCO to ADE20K and \textbf{+5.2\%} on COCO to UVO.

Our contributions in this paper include: 
\begin{itemize}
    \item A novel approach, {\bf G}eneric {\bf G}rouping {\bf N}etworks (\Ours{}), for open-world instance segmentation; GGN exploits additional pseudo ground truth supervision generated from learned pixel-level pairwise affinities.
    \item Comprehensive ablation experiments which provide insights about \Ours{} and the problem of open-world instance segmentation.
    \item \Ours{} achieve state-of-the-art performance in open-world instance segmentation on various benchmarks including COCO, LVIS, ADE20K, and UVO.
\end{itemize}

\section{Related works}
\noindent {\bf Object and instance segmentation}.
Before the success of deep learning, object segmentation approaches typically worked by grouping local regions into whole objects. Popular approaches include graph-based~\cite{gb,GBH}, Normalized Cut~\cite{868688}, Graph Cut~\cite{BoykovVZ01}, Multiscale Combinatorial Grouping~\cite{ArbelaezPBMM14}, and Selective Search~\cite{SelectiveSearch}. Since deep learning, end-to-end approaches proved their success on problems such as semantic segmentation~\cite{fcn,swin_transformer}, instance segmentation~\cite{8237584}, panoptic segmentation~\cite{KirillovHGRD19,MaX-DeepLab}. Despite sharing the common problem of segmentation, our approach is different in the open-world setup: instead of assuming a closed-world taxonomy, our work aims at detecting and segmenting both seen and unseen objects.

\noindent {\bf Pairwise affinity based approaches}.
Pairwise affinity is used in most graph-based segmentation methods~\cite{BoykovVZ01,868688,HeZC04} as an important term defining a relation graph of pixels for segmentation. The pixel-level pairwise affinity can be either hand-constructed~\cite{BoykovVZ01,868688,HeZC04} or learned~\cite{FowlkesMM03,KimLL13,TuragaBHDS09,Maire_2016_CVPR,LiuMGZ0K17,Liu2018AffinityDA,Gao2019SSAPSI}. Similar to pairwise affinity, object boundary detection~\cite{7410521,COB} is a dual problem but offers weaker supervision (sec~\ref{sec:experiments}) and cannot be trained as-is on non-exhaustive annotations. Different from previous approaches, instead of directly using the learned pairwise affinity for segmentation, we use it as an intermediate representation for pseudo-ground-truth generation which is later used to train our generic grouping model. Another difference between our approach and previous learned pairwise affinity comes from the open-world setting of the problem.

\noindent {\bf Open-world benchmarks and approaches}. Open-world setup~\cite{DeepMask15,bendale2015towards} has been recently (re)-introduced into various problems in computer vision such as recognition~\cite{liu2019large,Kong_2021_ICCV}, tracking~\cite{abs-2104-11221}, detection~\cite{Wang2020WhatLT,oln,joseph2021open}, and segmentation~\cite{uvo}. Among these, our work is mostly related to UVO~\cite{uvo} and OLN~\cite{oln}. We share the same problem of interest of open-world instance segmentation with UVO~\cite{uvo}. However, \cite{uvo} provides a new benchmark for the problem while our work provides an approach. Compared with the concurrent OLN work~\cite{oln}, which uses an objectness-based loss for generalization to unseen classes, our approach addresses generalization by combining pixel-level pairwise affinity with local grouping. Our work and OLN are orthogonal and complementary; as shown later, our approach alone is on par with OLN, and produces $4.5-5.7\%$ improvements when combined with OLN (see Table~\ref{tab:compare_oln} in section~\ref{sec:experiments}).

\section{Learning pairwise pixel affinities}
\label{sec:tech_pa}

\begin{figure}
  \centering
    \includegraphics[width=0.86\linewidth]{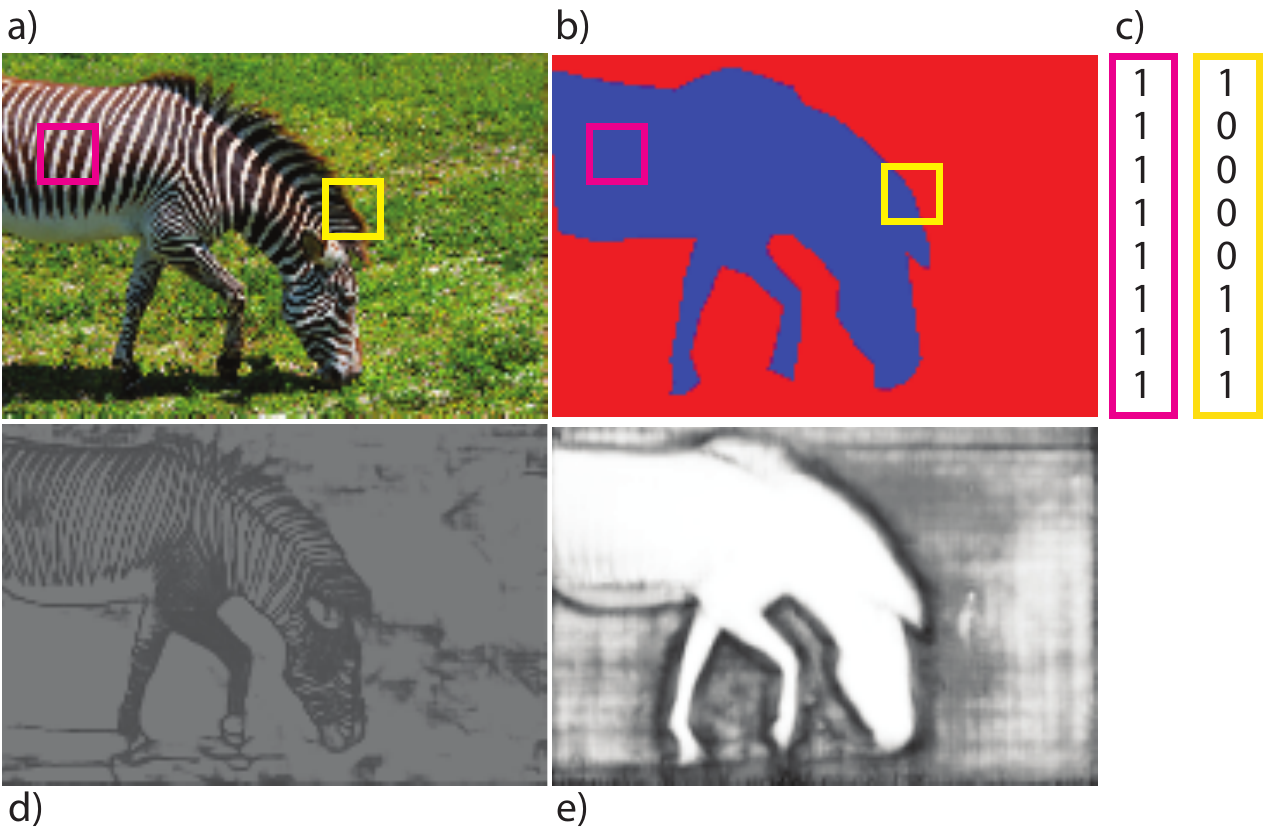}
  \vspace{-8pt}
  \caption{{\bf Pairwise affinity encoding and prediction}. Visualization of an example input image (a) and its corresponding ground truth mask annotation (b) and two pairwise-affinity encoding vectors (c) of the two center pixels of the two image patches marked with pink and yellow squares. The pink-patch's center pixel belongs to the same instance with all of its 8-neighbor pixels, thus it is encoded with a vector of all ones. The yellow-patch's center pixel lies at the object boundary and has 4 neighbor pixels belonging to background, thus it is encoded with a binary vector with four 0s and four 1s. (d) Edge prediction of the image in (a) using off-the-shelf edge detector~\cite{soria2020dexined}. (e) Pairwise affinity prediction of the image in (a) using our Pairwise Affinity predictor. Our predictor is trained using only person category masks from COCO. Best viewed in color.}
  \label{fig:pa_encoding}
  \vspace{-10pt}
\end{figure}

Grouping can be locally represented by pixel pairwise relationship: whether two neighboring pixels should be grouped together or not. Given a 3-channel RGB input image $I \in \mathbb{R}^{3 \times H \times W}$, we consider a pixel's pairwise relationship in a $3\times3$ neighborhood. This gives a pairwise affinities map of $P \in \{0,1\}^{8 \times H \times W}$ and $P_{i,j} \in \{0,1\}^8$ encodes the local pixel-level pairwise affinities of the pixel $(i,j)$ with its 8-neighboring pixels in the image $I$. Figure~\ref{fig:pa_encoding}~(c) illustrates the pairwise affinity encodings of the two pixels at the centers of the two image patches marked with the pink and yellow squares in Figure~\ref{fig:pa_encoding}~(a) and (b). We use pixel-level-prediction convolutional neural networks to predict $P$ (\emph{PA Pred.}, Figure~\ref{fig:overview} a), such as FCN~\cite{fcn} and UPerNet~\cite{UPerNet}. We remark that this is a dual problem to binary object boundary detection (\cite{7410521,COB}), and techniques used for training \emph{PA Pred.} can also be adopted to binary object boundary detectors to serve as the local representation for our framework.

\noindent {\bf Training from non-exhaustive segmentation masks}. Ideally, if all pixels in the image are exhaustively annotated with instance segmentation masks, e.g., all object boundaries are labelled, then all pairs of neighboring pixels can provide good supervision signal for learning pairwise affinity. However, exhaustive annotations for instance segmentation are expensive and time-consuming to obtain, so most datasets come with non-exhaustive segmentation masks (eg. COCO~\cite{Lin2014MicrosoftCC}). In particular, non-annotated out-of-taxonomy objects cannot be distinguished from background pixels. To address this, we only use neighboring pixels with object--object or object--background relations for training pairwise affinities; we ignore the unreliable background-background pairs. In addition, training of pairwise affinities is unbalanced: only pixels at an object's boundary has zero-valued affinities; all other pixels have affinity 1 to all their neighbors. We weight the positive affinities by computing the ratio between the positive affinities and negative affinities on a subset of training data (e.g., 0.05 for positive).



\begin{figure}
\centering
{\small 
 \begin{subfigure}[b]{0.32\linewidth}
    {\includegraphics[width=\linewidth]{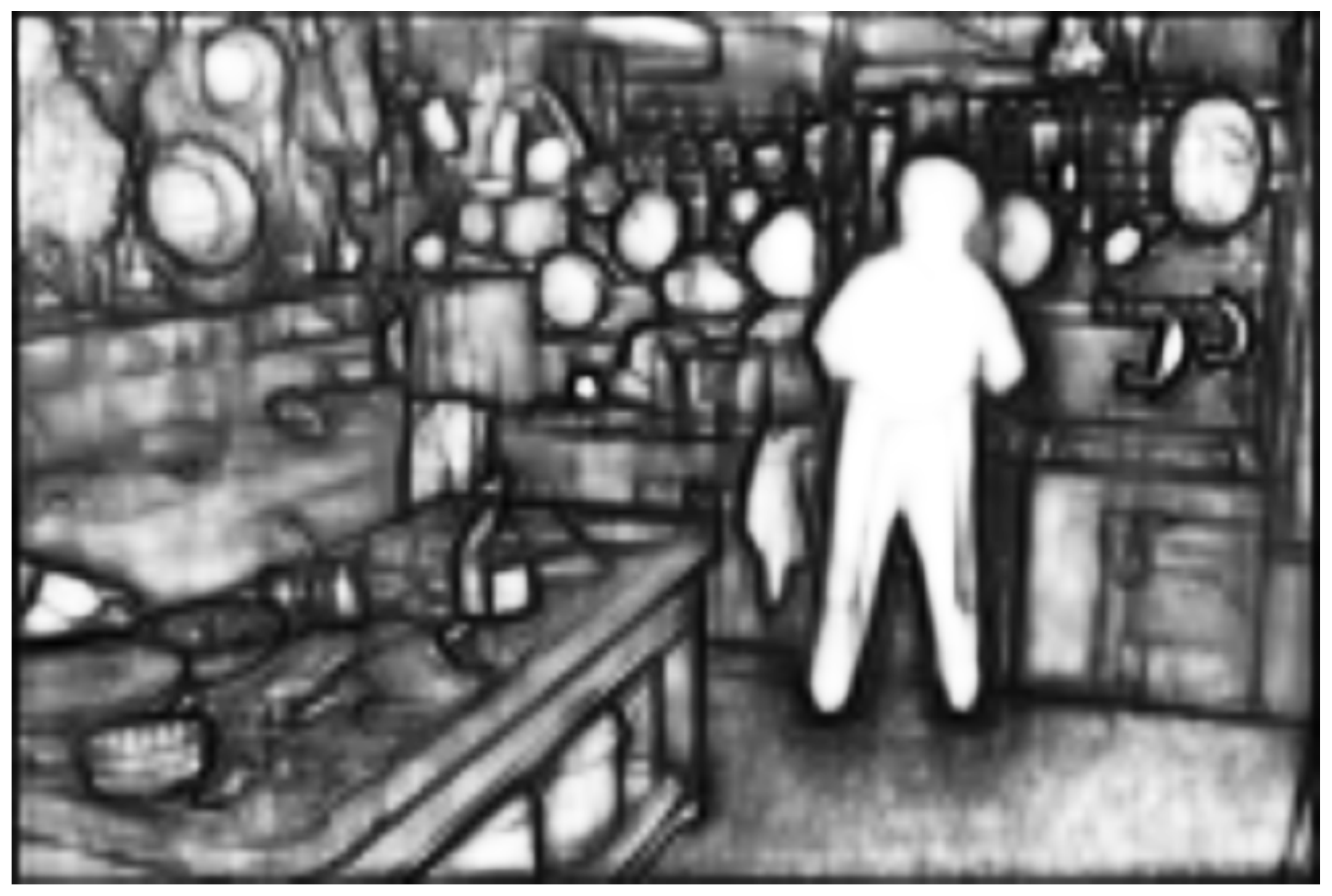}}
 \end{subfigure}
 \begin{subfigure}[b]{0.29\linewidth}
    {\includegraphics[width=\linewidth]{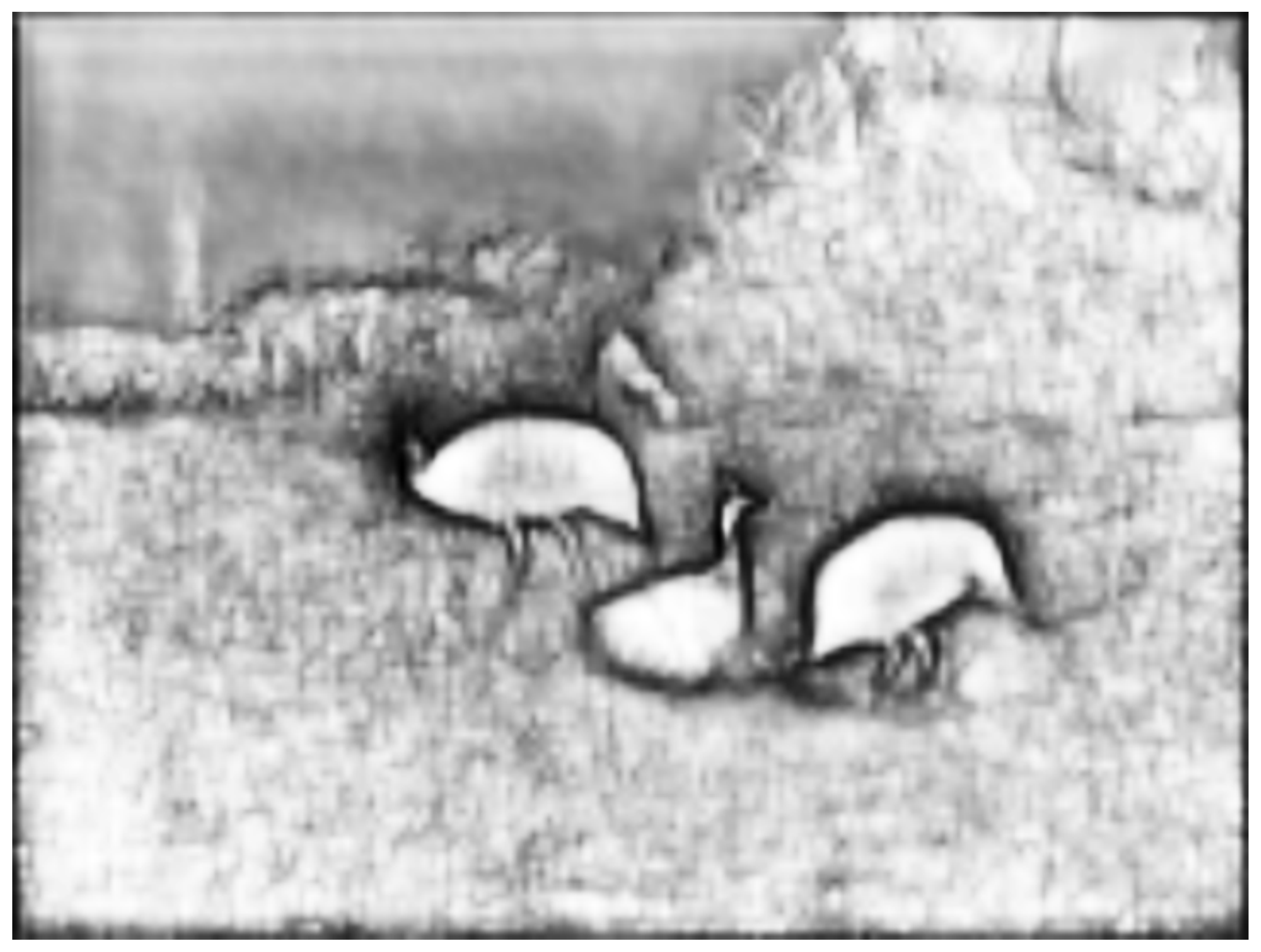}}
 \end{subfigure}
 \begin{subfigure}[b]{0.32\linewidth}
    {\includegraphics[width=\linewidth]{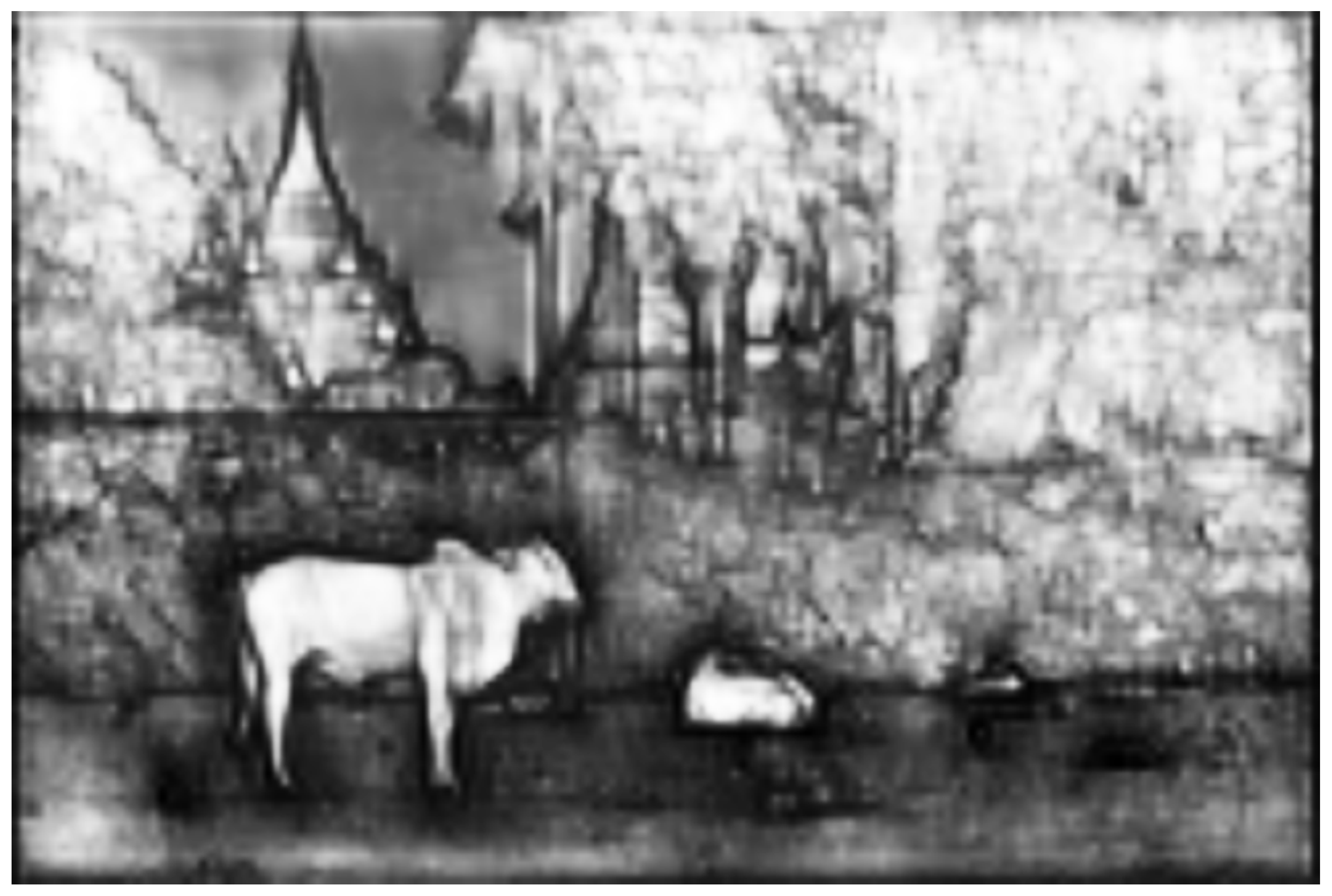}}
 \end{subfigure}
 \begin{subfigure}[b]{0.32\linewidth}
    {\includegraphics[width=\linewidth]{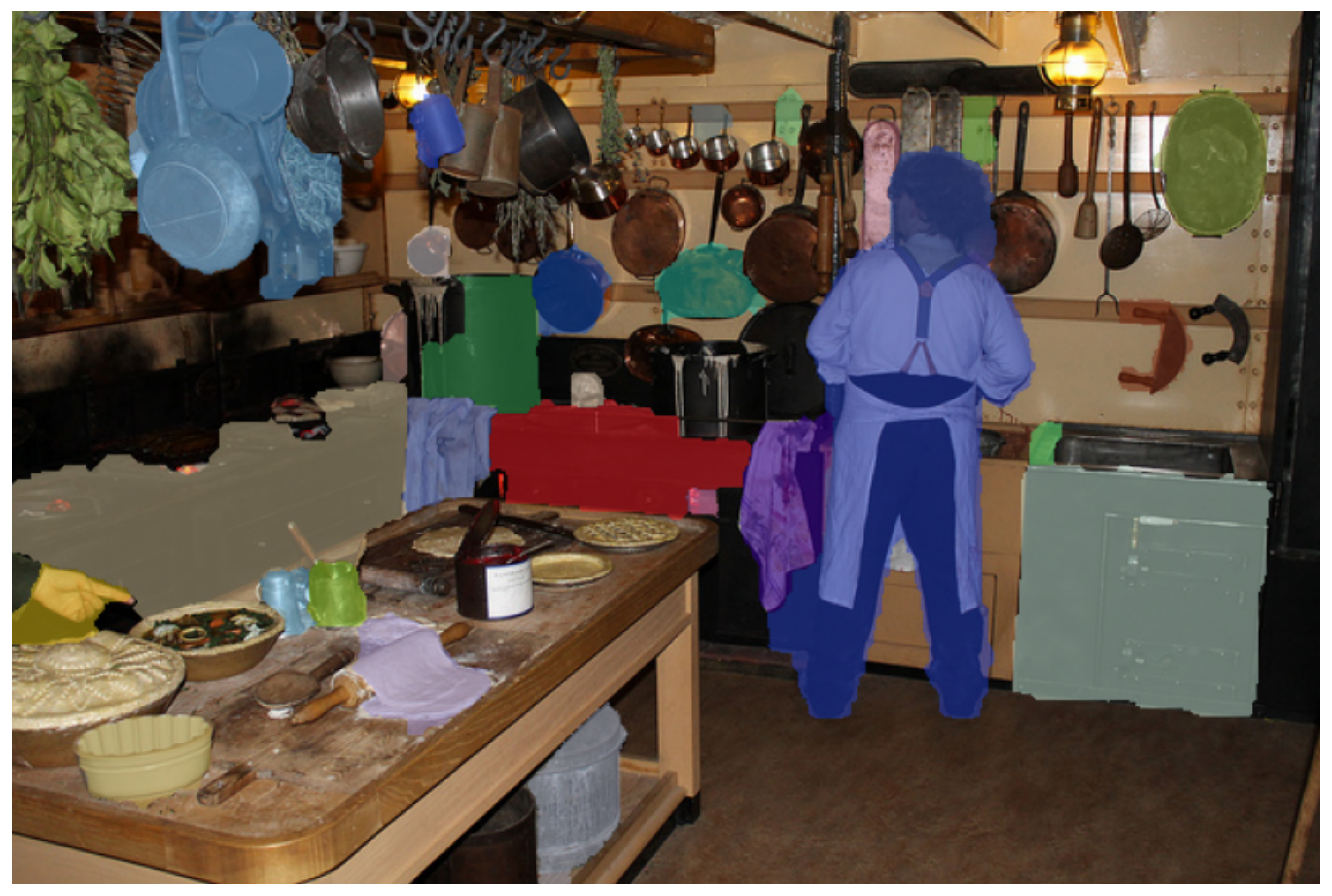}}
 \end{subfigure}
 \begin{subfigure}[b]{0.29\linewidth}
    {\includegraphics[width=\linewidth]{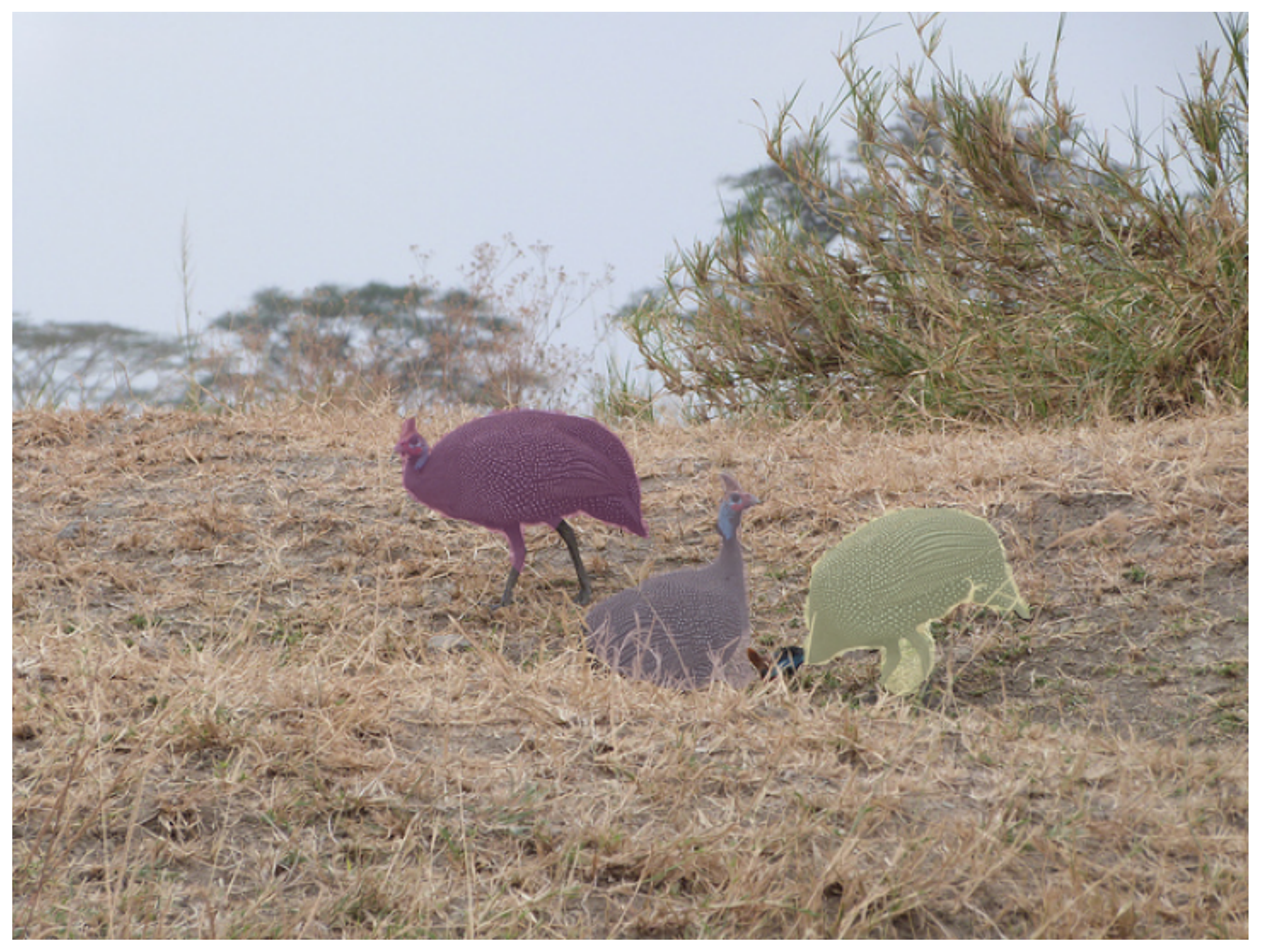}}
 \end{subfigure}
 \begin{subfigure}[b]{0.32\linewidth}
    {\includegraphics[width=\linewidth]{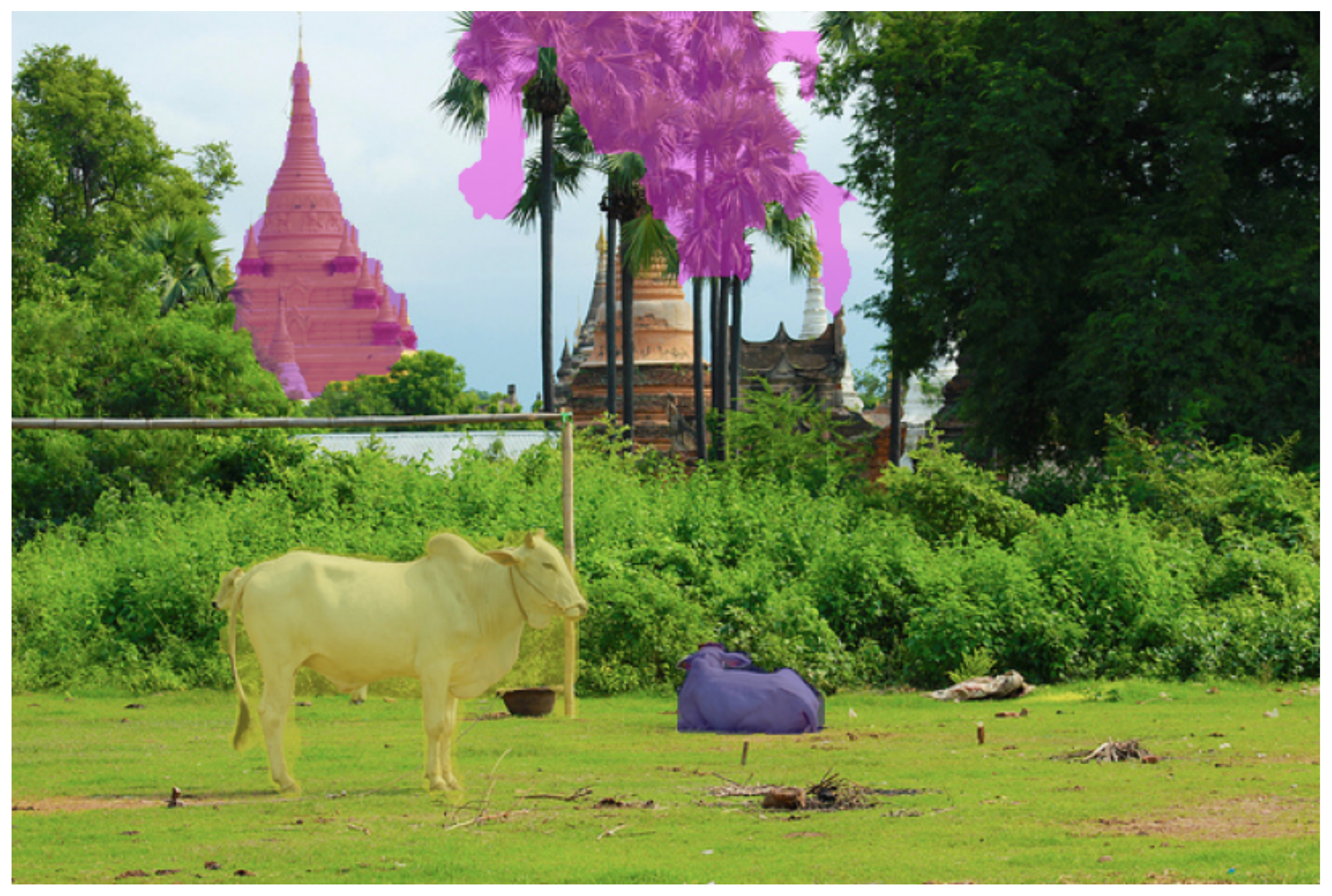}}
 \end{subfigure}
 \vspace{-8pt}
 \caption{{\bf Visualization of predicted pairwise affinities and generated pseudo masks, trained only on Person class in COCO.} Despite only seeing masks of Person during training, the PA predictor correctly captures pairwise relationships of other types of objects (top-row). By grouping pixels based on predicted PA, we can generate pseudo masks of other categories (bottom-row).
 }
 \label{fig:vis_pa_mask}
 }
 \vspace{-8pt}
\end{figure}

\noindent {\bf What does a Pairwise Affinity predictor learn?}
Intuitively, an ideal PA measure should discriminate between instance boundaries and instance interiors: i.e. whether two neighboring pixels cross an object boundary or not. Ideally, our PA predictor should be robust to boundaries of all objects, not just the categories seen during training; this is a key requirement for success in the open-world setting. Indeed, this is the case: our PA predictor learns instance boundaries and generalizes well to unseen classes. Figure~\ref{fig:vis_pa_mask} top-row and Figure~\ref{fig:pa_encoding}~(e) visualize pairwise affinity predictions from our PA predictor, trained using ground truth masks of \textbf{only} the person category from COCO. Our PA predictor learns to generalize to unseen classes such as zebra, bird, temple, and cooking pan. We note that `person' masks are particularly diverse owing to clothing and accessories. This is helpful for models to learn and generalize the notion of semantic boundaries. We show quantitative results on PA's generalization in section~\ref{sec:pa_ablation} and exploit this generalization behavior to improve segmentation in section~\ref{sec:tech_aug_pseudo}. Finally, we point out that pairwise affinity should capture the semantics of instance boundaries; this is quite different from visual edge maps because many visual edges are not instance boundaries. Figure~\ref{fig:pa_encoding}~(d) presents the edge prediction by an off-the-shelf edge detector~\cite{soria2020dexined}. Many visual edges on the back of the zebra are clearly not object boundaries, but are still detected by an edge detector.

\section{Augmenting with pseudo ground truth} \label{sec:tech_aug_pseudo}

Existing state-of-the-art detectors and instance segmentation models, such as Mask R-CNN, often fail to detect and segment novel objects unseen during training. This can be caused by difficulties to group pixels into unknown entities due to lack of supervision signals during training. In addition, even a novel region is grouped and proposed, a generic concept of objectness is missing and such out-taxonomy proposals are suppressed. We kill two birds with one stone, by using pseudo-GT masks generated from PA to train these detectors. The pseudo-masks benefit from pixel diversity to provide novel segments not seen during training, and therefore enhance supervision signals for both novel grouping and a more inclusive concept of objectness.

\noindent {\bf Grouping pixels into regions}. Based on predicted pairwise affinities, we leverage class-agnostic local grouping algorithms to group pixels into instances. One may use the \emph{Connected Component} (CC) algorithm for grouping. CC treats all affinities independently using a hard cut-off threshold to decide pixel connections which may be a sensitive parameter to tune. Alternatively one may use \emph{graph-based hierarchical} grouping (GBH)~\cite{gb} which is a variant of agglomerative clustering. In segmentation literature~\cite{ArbelaezMFM11,ZhangM20}, Oriented Watershed Transform (OWT), globalized contour through Normalized Cut (gPb)~\cite{ShiM00} and Ultrametric Contour Map (UCM)\cite{ucm} are also used for grouping from the edge map of an image. Following~\cite{ucm,ArbelaezPBMM14}, we first aggregate the image pairwise affinity map into a semantic edge map using pooling along the channel dimension, e.g., reducing from 8-channels to 1-channel. This semantic edge map is passed to OWT to generate initial segments, whose edges are then globalized through normalized cut. We take the average of semantic edge map and its globalized version as input to UCM for grouping. We acknowledge that a different linear combination of these two might work better upon further study. We provide the ablation comparing CC, GBH, and different components of OWT+gPb+UCM in section~\ref{sec:experiments}.

\noindent {\bf Computing objectness}. Objectness~\cite{EndresH10} measures grouping qualities; in our framework, it is critical to decide which pseudo-GT masks to select for training detectors. An ideal objectness score should reveal over-segmentation and under-segmentation. In previous literature, objectness can be modeled by low-level features such as shape and contours such as MCG~\cite{ArbelaezPBMM14} or learned directly from annotated data as classification (Region Proposal Network~\cite{FasterRCNN}) or regression (Object Localization Network~\cite{oln}). We consider both types of objectness. For low-level features, we use predicted pairwise affinities to define objectness score of each region $R$ by total affinities $\mathcal{O}_{PA}(R)$:
\begin{equation}
\mathcal{O}_{PA}(R) = \frac{Inner(R)}{R_{inner}}-\frac{Outer(R)}{R_{boundary}}
\label{equ:objectness_pa}
\end{equation}
where $Inner(R)$ and $Outer(R)$ are the inner and outer affinities of $R$ defined by the sum of pairwise affinities of pixels inside or crossing-boundary of $R$, respectively. $R_{inner}$ and $R_{boundary}$ denote the number of pixels inside $R$ and on the boundary of $R$. Intuitively, we want to rank high for the region with strong inner pairwise affinities and weak affinities at the boundary (a.k.a strong cut). For learned objectness, we consider scoring from OLN~\cite{oln} $\mathcal{O}_{OLN}(R)$ :
\begin{equation}
\mathcal{O}_{OLN}(R) = \sqrt{centerness(R) * IoUness(R)}
\label{equ:objectness_oln}
\end{equation}
where $centerness(R)$ and $IoUness(R)$ are the centerness and IoU predictions of the bounding box of $R$. We can optionally combine $\mathcal{O}_{PA}$ and $\mathcal{O}_{OLN}$ by taking average.

\noindent {\bf Generic Grouping Networks (\Ours{})}. We generate class-agnostic masks from PA predictor and grouping module and use the objectness score to rank regions provided by grouping methods (Figure~\ref{fig:overview} b). We then select top ranked regions from each image as pseudo ground-truth (GT) masks for training our generic object proposal network (Figure~\ref{fig:overview} c). Since the whole approach to generate pseudo-GT masks are designed in a class-agnostic grouping fashion, we expect the pseudo-GT masks to cover a diverse set of objects and parts, and more importantly most of them are from unseen categories, as shown in Figure~\ref{fig:vis_pa_mask} bottom-row. Since our \Our{} is trained on a large and diverse set of masks, it is expected to generalize to unseen classes, thus providing a good solution for open-world instance segmentation. \Our{} is generic in the sense of both pixels and models: it can work on different domains of images, labeled or unlabeled, and it can work on any architecture for object detection or segmentation, such as Faster R-CNN~\cite{FasterRCNN}, Mask R-CNN~\cite{8237584}, YOLO~\cite{7780460}, or Swin Transformer~\cite{swin_transformer}. The adoption is as simple as making the multi-class classification prediction head into a binary foreground vs. background classification head. 


\section{Experiments}
\label{sec:experiments}

\subsection{Implementation Details}
\noindent {\bf Datasets}. We conduct experiments on COCO17~\cite{Lin2014MicrosoftCC}, LVIS~\cite{8954457}, ADE20K~\cite{8100027}, and UVO~\cite{uvo}. \textbf{COCO} is a standard benchmark for instance segmentation with $80$ object categories annotated on 164k images. \textbf{LVIS} is an instance segmentation dataset with $1203$ classes in a long-tail distribution. It is labeled as a federated dataset and does not include exhaustive label for its categories. We adopt LVIS to study cross-category generalization when a large taxonomy is provided. \textbf{ADE20k} is a semantic segmentation dataset with all pixels exhaustively annotated by object instances or stuff. \textbf{UVO} is a video instance segmentation dataset of YouTube videos (Kinetics400~\cite{kinetics}) with object masks exhaustively labeled. We use validation set of ADE20K (2000 images) and UVO sparse (7356 frames) to evaluate open-world segmentation in the wild in section~\ref{sec:wild}. In all setups, we use only mask annotation (without class labels) for training and evaluation in the open-world, class-agnostic. We note that PA predictor, baseline methods (e.g., Mask R-CNN), and \Our{} has access to the \textbf{same} labeled masks.

\noindent {\bf Backbone architectures and loss function}. We adopt UperNet~\cite{UPerNet} for our PA predictor to learn pairwise affinities. For training, our generic grouping networks, we use Mask R-CNN~\cite{8237584} with a ResNet-50 backbone as a default setup. Unless specified otherwise, models are initialized by ImageNet~\cite{deng2009imagenet} pre-training. We use Binary Cross Entropy loss to train pairwise affinities. We ignore background-background affinity as in section~\ref{sec:tech_pa}. We note that back-propagating losses that include background-background affinities leads to very poor cross-category generalization (e.g., $-15\%$ Average Recall).

\noindent {\bf Ranking and selecting pseudo-GT masks}. Unless otherwise specified, we use $\mathcal{O}_{PA}$ (Eq.~\ref{equ:objectness_pa}) to rank pseudo-GT masks. We pick top-k pseudo-GT masks per image (k$\in [1,3]$), where k is selected to improve unseen categories performance while minimally impact seen performance.

\noindent {\bf Training and evaluation}. We build model training and inference on MMDet~\cite{mmdetection} platform; all training are done with the default 1x schedule. Following previous object proposal literature~\cite{SelectiveSearch,DeepMask15}, we use \textbf{average recall} (AR) over multiple IoU thresholds (0.5:0.95) to evaluate model performance.


\noindent {\bf Cross-category evaluation}. Cross-category generalization is a major challenge for open-world: how do we detect and segment objects whose categories are outside training data. We split existing datasets by their categories to construct controlled environment for ablations (Table~\ref{tab:datset_split}). In each setup, we train PA and baseline methods with the same splits of categories (no additional supervision for PA).

On COCO dataset, we follow common practice~\cite{DeepMask15,oln} to split COCO into $20$ classes overlapped with Pascal VOC~\cite{PascalVOC} for training (seen) and use the rest of $60$ COCO-exclusive classes for evaluation (unseen). We further include an extreme case by using only \textit{person} class for training and the rest $79$ classes for evaluation. 

On LVIS dataset, some categories are highly overlapped: for example, clothes (``jacket'', ``wet suit'') are highly overlapped with ``person'' when the person is wearing the clothes. In a class-agnostic setup, a detector trained with person masks can detect clothes as person and vice versa. Other examples of high-overlapped category pairs are ``ball'' with ``tennis ball'', ``alcohol'' with ``beer bottle'', or ``computer monitor'' with ``television set''. COCO and LVIS share the same set of images, but with different annotations. LVIS covers $1203$ categories which include all $80$ categories from COCO. COCO also exhaustively annotates all masks of objects that belong to its 80 categories while LVIS is annotated so as to maintain a similar number of masks across categories. This means that some object instances, even if they belong to LVIS categories, are not annotated. As COCO masks are more exhaustively annotated, we use COCO masks and validate cross-category overlap with LVIS masks. We find that there are 67k LVIS masks outside COCO taxonomy having $>$0.5 IoU with COCO masks. To ensure a clear distinction between seen and unseen categories, we remove those masks in both training and validation for cross-category generalization evaluation. We study transfer performance by training on COCO categories and evaluate on LVIS non-COCO categories and vice-versa.

\begin{table}
  \centering
  {\small
  \begin{tabular}{|c|c|c|c|c|}
  \hline
    Dataset & Train & Eval & Image & Mask \\
    \hline
    \multirow{2}{*}{COCO} & Person(1) & non-Person & 64k & 161k \\
     & VOC(20) & non-VOC & 95k & 493k \\
    \hline
    \multirow{3}{*}{LVIS} & COCO(80) & non-COCO & 100k & 455k \\
     & non-COCO(1122) & COCO & 85k & 749k \\
     & +Person(1123) & non-Person & 86k & 775k \\
  \hline
  \end{tabular}}
  \vspace{-8pt}
  \caption{\textbf{Cross-category generalization evaluation setups.} We split categories in COCO and LVIS to evaluate.}
  \label{tab:datset_split}
\end{table}

\subsection{Learning Pairwise Affinities: Ablation Study} \label{sec:pa_ablation}

\begin{table}
  \centering
  {\small
  \begin{tabular}{|c|c|c|c|c|c|}
    \hline
    Grouping & CC & GBH & WT+UCM & +OWT & +gPb \\
    \hline
    Recall@all & 14.4 & 17.1 & 23.6 & 23.8 & \textbf{24.2} \\
    \hline
  \end{tabular}}
  \vspace{-8pt}
  \caption{{{\bf Comparing different grouping methods}. Methods are applied on the same affinity maps and output roughly a similar number of proposals. OWT+gPb+UCM gives the best recall.}}
  \label{tab:compare_grouping}
\end{table}

\begin{table}
  \centering
  {\small
  \begin{tabular}{|c|c|c|c|}
    \hline
    Aggregate & Min & Max & Mean  \\
    \hline
    8-channel & \underline{22.8}/\textbf{19.3} & 18.1/16.7 & 22.1/\underline{18.9} \\
    1-channel & 19.9/18.4 & NA/NA & \textbf{23.1}/18.5 \\ 
    \hline
  \end{tabular}}
  \vspace{-8pt}
  \caption{\textbf{The effect of different PA aggregation.} Evaluated by AR@100 on both seen and unseen categories (VOC/non-VOC) and results are separated by /. 1-channel prediction is not applicable with max-pooling since the prediction target is all 1s (all pixels have at least one neighbor connected). 8-channel PA prediction with min pooling provides the best AR.} 
  \label{tab:compare_target}
\end{table}

\noindent \textbf{Grouping mechanisms.} We revisit different grouping methods to construct segment masks from pairwise affinities: Connected Component (CC), Graph-Based Hierarchical~\cite{GBH} (GBH) and methods based on Ultrametric Contour Map~\cite{ucm} (UCM). In UCM, we ablate the effect of orientation in watershed transform (OWT vs. WT), and the effect of including globalized edge (gPb). To evaluate, we generate mask outputs from each of the method and directly evaluate their AR. We tune parameters for each method so that each has roughly the same number of output segments on average. Since CC gives a single non-overlapping output instead of a hierarchical structure like GBH or UCM, we use multiple thresholds and use all segments from each threshold. We found that UCM-based methods significantly outperforms other two methods (Table~\ref{tab:compare_grouping}): whereas CC and GBH make decision on a merge based on a single pairwise relationship, UCM uses all relations between two segments, and is more robust. In addition, adding orientation and gPb further improves grouping results.

\noindent \textbf{PA aggregation.} In UCM, PA (8 neighbors) need to be aggregated into one value to feed to WT. The aggregation can be implemented by a pooling operation and can be applied before or after PA prediction. Specifically, we can either: (i) train a PA predictor to predict a 8-channel PA map then apply aggregation on the prediction output of the PA predictor; or (ii) train a PA predictor to predict a 1-channel PA map which is the aggregated version of ground truth. We compare different methods for aggregating pairwise affinity values (Table~\ref{tab:compare_rep}) and found min aggregation works the best. Alternatively, we can directly predict the aggregated pairwise affinities. We found that a single-channel prediction of mean of pairwise affinities works comparably with 8-neighbor predictions (Table~\ref{tab:compare_rep}). We remark that ``1-channel, min'' is equivalent to adopting a binary boundary detector trainer in our framework (e.g., HED~\cite{7410521}), which offers weaker supervision signals than PA.


\subsection{Cross-category evaluation of \Our{}}
\label{sec:cross_category}

We use the pseudo-GT masks from PA+Grouping to train detectors for open-world segmentation. We optionally use additional ground truth masks when they are available. Since PA generalize well in the open-world, the pseudo masks offer more diversities to the training data and therefore improve generalization of downstream detectors (\Ours{}). We begin by comparing PA with other candidate representations for open-world segmentation.

\noindent \textbf{Pairwise affinity is a strong representation for open-world.} Besides pairwise affinities, we consider a few other types of mid-level representations to encode grouping and generalize in the open-world:
\begin{itemize}
    \item \textbf{Edge maps} are strong alternative to PA to encode grouping. We take SOTA edge detector DexiNed~\cite{soria2020dexined}. 
    \item \textbf{Depth maps} by pretrained Mannequin network~\cite{li2019learning}. 
    \item \textbf{Feature affinities} computed on semantic features, self-supervised trained on ImageNet (MoCoV2~\cite{chen2020mocov2}).
\end{itemize}

Most of the features here, except edge map, are not proper to run UCM to construct grouping. Therefore, we consider replacing the RGB input with the proposed representation (e.g., depth map or PA) to understand how well the representation can generalize in the open-world compared to RGB in cross-category evaluation. 

Surprisingly, all representations have regularizing effects compared to RGB to improve generalization to unseen classes when only training on person (Table~\ref{tab:compare_rep}). Pairwise affinities outperform all other types of representations regardless of application methodologies (replacing input or adding pseudo masks). In particular, using UCM to generate pseudo-GT masks for edge does not benefit much, since without semantics, edge map can over-segment entities.


\begin{table}
  \centering
  {\small
  \begin{tabular}{|c|c|c|c|c|c|c|c|}
    \hline
    \multirow{2}{*}{\footnotesize Method} & \multicolumn{5}{c|}{replace RGB} & \multicolumn{2}{c|}{UCM mask} \\
    \cline{2-8}
     & \footnotesize RGB & \footnotesize depth & \footnotesize edge & \footnotesize MoCo & \footnotesize PA & \footnotesize edge & \footnotesize PA \\
    \hline
    \scriptsize{nonPerson} & 4.9 & 10.9 & 10.5 & 10.7 & \underline{14.1} & 7.9 & \textbf{20.9} \\
    \scriptsize{nonVOC} & 19.9 & 17.8 & 21.3 & 21.8 & \underline{26.5} & 19.7 & \textbf{28.7} \\
    \hline
  \end{tabular}}
  \vspace{-8pt}
  \caption{\textbf{Compare Pairwise Affinity with other types of mid-level representations}. All mid-level representations can serve to help generalizing to unseen categories in certain scenario to certain degrees. The significant improvement of pairwise affinities over edge map shows the importance for boundaries to contain semantics. Methods are evaluated by AR@100. Pairwise Affinities provide strongest generalization signals for open-world grouping.}
  \label{tab:compare_rep}
\end{table}

\noindent \textbf{\Our{} significantly outperforms baselines on cross-category generalization.} We take top-scoring pseudo-GT masks and use them in addition with the ground truth masks; we remove pseudo-GT masks whose IoU overlap with in-taxonomy GT masks are greater than 0.5. 

\Our{} has significantly stronger cross-category generalization compared to baseline Mask R-CNN (Table~\ref{tab:diff_setup}). On low to medium-sized taxonomy on COCO dataset, \Our{} achieves \textbf{+16\%} AR@100 gain and \textbf{+8.8\%} AR@100 gain when training on Person-only and training on 20 VOC classes, respectively. In the large-taxonomy setup, \Our{} achieves 1.8\% to 3.9\% gain on AR@100 in different setups. The gain is smaller when training on non-COCO categories; we believe that this is caused by the fine-grained taxonomy of LVIS: many classes in LVIS are objects parts or parts of other classes. Training on non-COCO categories on LVIS makes pairwise affinities closer to edge maps.

Additionally, we evaluate pseudo-GT masks generated by pairwise affinities with UCM. This is equivalent to our ~\Our{} but without using top-down instance-level training (as in Figure~\ref{fig:overview} b), denoted as \emph{PA+Grouping}. Comparing with Mask R-CNN baseline, local grouping using learned pairwise affinities offer stronger performance in low to medium sized taxonomy. Finally, \Our{} significantly outperforms the \emph{PA+Grouping} baseline which suggests the benefits of instance-level end-to-end training on pseudo-GT masks.

\begin{table}
  \centering
  {\small
  \begin{tabular}{|c|c|c|c|c|}
  \hline
    \multirow{2}{*}{Train (\# classes)} & Mask & PA+ &  \multirow{2}{*}{\Our{}} & \textcolor{gray}{Upper} \\
     & R-CNN & Grouping &  & \textcolor{gray}{Bound} \\ 
    \hline 
    \hline
    \multicolumn{5}{|c|}{COCO-Dataset} \\
    \hline
    Person (1) & 4.9 & 14.6 & \textbf{20.9} & \textcolor{gray}{49.2} \\
    VOC (20) & 19.9 & 22.0 & \textbf{28.7} & \textcolor{gray}{49.6} \\
    \hline 
    \hline
    \multicolumn{5}{|c|}{LVIS-Dataset} \\
    \hline
    COCO (80) & 16.5 & 17.1 & \textbf{20.4} & \textcolor{gray}{36.1} \\
    non-COCO (1123) & 21.7 & 16.2 & \textbf{23.6} & \textcolor{gray}{35.1} \\
    +Person (1124) & 27.3 & 18.4 & \textbf{29.1} & \textcolor{gray}{44.2} \\
    \hline 
  \end{tabular}}
  \vspace{-8pt}
  \caption{\textbf{\Our{} generalizes to out-taxonomy categories significantly better than baselines.} \Our{} also outperforms the pseudo-GT masks generated by pairwise affinities (denoted as PA+Grouping), proving the benefit of the instance-level training with additional pseudo-GT supervision. Upper bound indicates AR@100 achieved by training on entire taxonomy (all classes). }
  \label{tab:diff_setup}
\end{table}

\noindent \textbf{\Our{} is comparable and complementary to state-of-the-arts object proposal method}. Object Localization Network (OLN)~\cite{oln} is a concurrent work to tackle open-world object proposal. OLN proposes to replace classification with localization quality prediction to avoid overfitting to annotated objects, which is similar to how we train pairwise affinities by not backpropagating the loss in un-annotated relationships. Different from OLN, pseudo masks generated bring more diversity to training data, and therefore help to generalize better. We compare \Our{} with OLN in Table~\ref{tab:compare_oln} and find that \Our{} achieves similar performance as OLN ($-1.5\%$ on box AR@100, $+1.8\%$ on mask AR@100). When adding $\mathcal{O}_{OLN}$ (Eq.~\ref{equ:objectness_oln}) to rank and select pseudo-GT masks, \Our{} improves by 2.2\%. When adopting OLN as backbone, \Our{} sets new state-of-the-arts for cross-category generalization of VOC to COCO.

\begin{table}
  \centering
  {\small
  \begin{tabular}{|c|c|c|c|c|c|}
    \hline
    \multirow{2}{*}{Backbone} & \multirow{2}{*}{Base} & \multirow{2}{*}{OLN} & \multirow{2}{*}{\Our{}} & \Our{}+ & \Our{}+  \\
     & & & & $\mathcal{O}_{OLN}$ & OLN \\
    \hline
    Faster R-CNN & 24.9 & 33.0 & 31.5 & 34.7 & \textbf{37.2} \\
    Mask R-CNN & 19.9 & 26.9 & 28.7 & 30.9 & \textbf{33.7} \\
    \hline
  \end{tabular}}
  \vspace{-8pt}
  \caption{\textbf{\Our{} is competitive yet complementary to OLN.} Trained on VOC and evaluated on non-VOC using AR@100. Adopting $\mathcal{O}_{OLN}$ improves ranking of pseudo-masks and thus improves \Our{}; adopting OLN backbone further improves.}
  \label{tab:compare_oln}
\end{table}


\subsection{Evaluate open-world segmentation in the wild}
\label{sec:wild}

\begin{table}
  \centering
  {\small
  \begin{tabular}{|c|c|c|c|c|c|}
    \hline
    \multirow{2}{*}{Method} & \multirow{2}{*}{Ranking} & \multicolumn{2}{c|}{ADE20K} & \multicolumn{2}{c|}{UVO} \\
    \cline{3-6}
     & & AR & AP & AR & AP \\
    \hline
    \multicolumn{2}{|c|}{\textcolor{gray}{Selective Search}} & \textcolor{gray}{3.8} & - & \textcolor{gray}{4.7} & - \\
    \multicolumn{2}{|c|}{Mask R-CNN} & 14.7 & 6.4 & 40.1 & 18.5 \\
    \hline
    \multirow{2}{*}{\Our{}} & $\mathcal{O}_{PA}$ & 18.3 & 7.9 & 42.6 & 19.4 \\
     & $\mathcal{O}_{PA}+\mathcal{O}_{OLN}$ & \underline{21.0} & \textbf{9.7} & \underline{43.4} & \underline{20.3} \\
    \hline
    \multicolumn{2}{|c|}{\Our{}, pseudo-GT pre-training} & \textbf{21.5} & \underline{9.3} & \textbf{45.3} & \textbf{21.0} \\
    \hline
  \end{tabular}}
  \vspace{-8pt}
  \caption{\textbf{Open-world segmentation in the wild on ADE20K and UVO.} \Our{} significantly outperforms the baseline Mask R-CNN when using the same amount of training data and annotation. In addition, having stronger objectness by combining $\mathcal{O}_{PA}$ and $\mathcal{O}_{OLN}$ further improves model performance. Finally, replacing ImageNet label pre-training with ImageNet pseudo-GT masks pre-training (sec~\ref{sec:scale_unlabeled}) offers additional improvement.}
  \label{tab:compare_wild}
\end{table}

\begin{figure*}[t]
\centering
{\small 
 \begin{subfigure}[b]{0.19\linewidth}
    {\includegraphics[width=\linewidth]{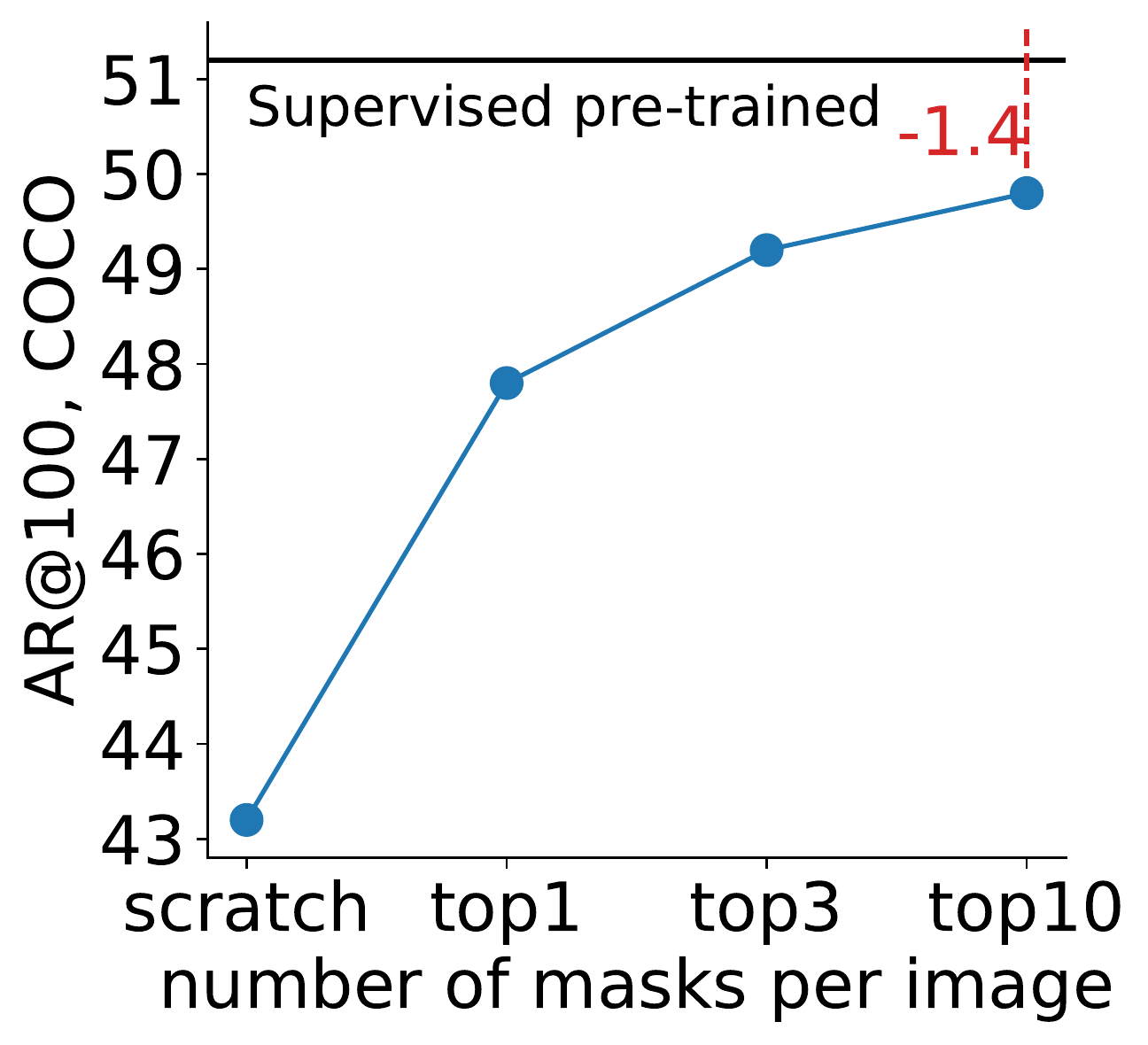}}
    \caption{COCO}
 \end{subfigure}
 \begin{subfigure}[b]{0.19\linewidth}
    {\includegraphics[width=\linewidth]{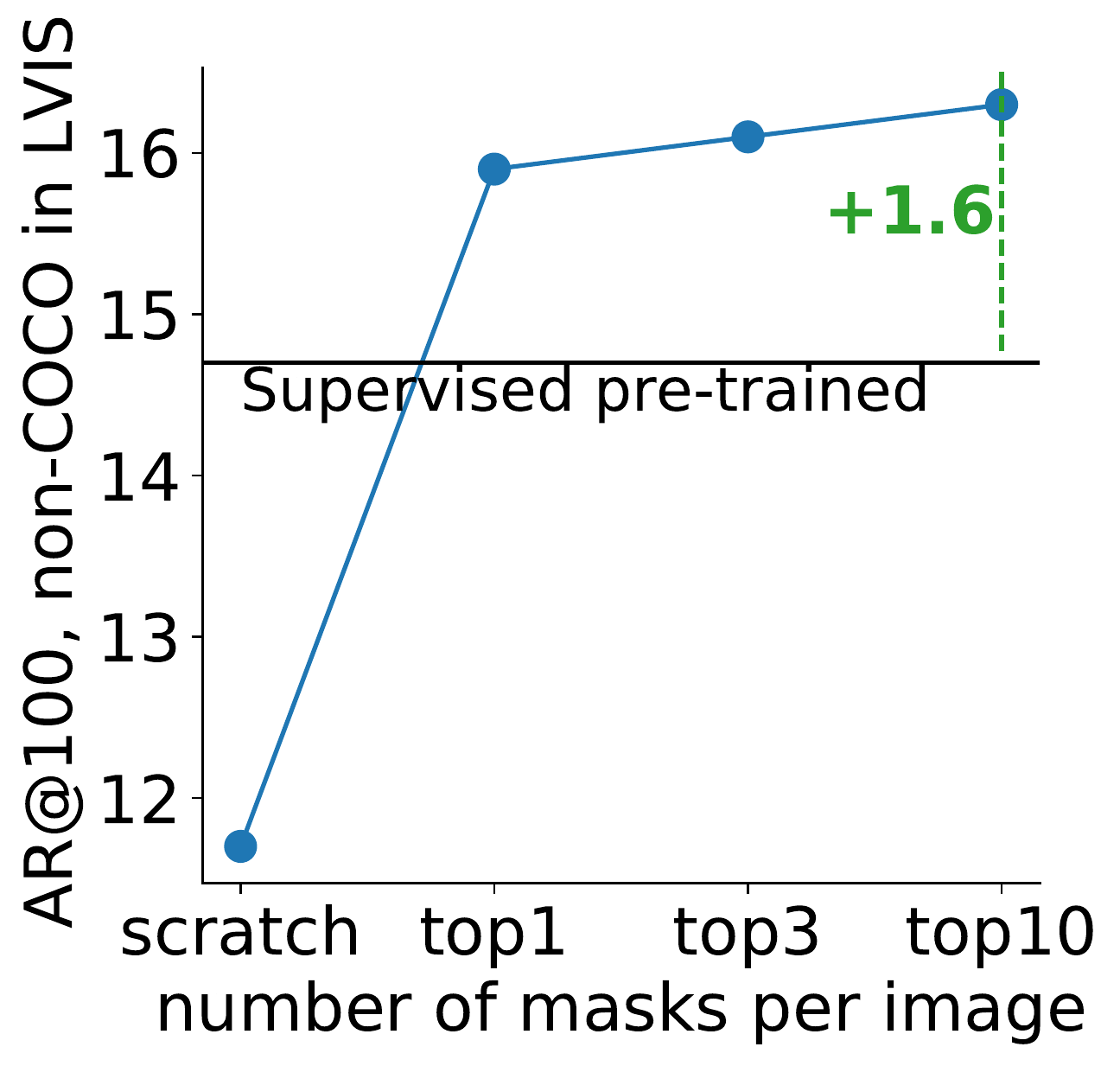}}
    \caption{LVIS, non-COCO}
 \end{subfigure}
 \begin{subfigure}[b]{0.19\linewidth}
    {\includegraphics[width=\linewidth]{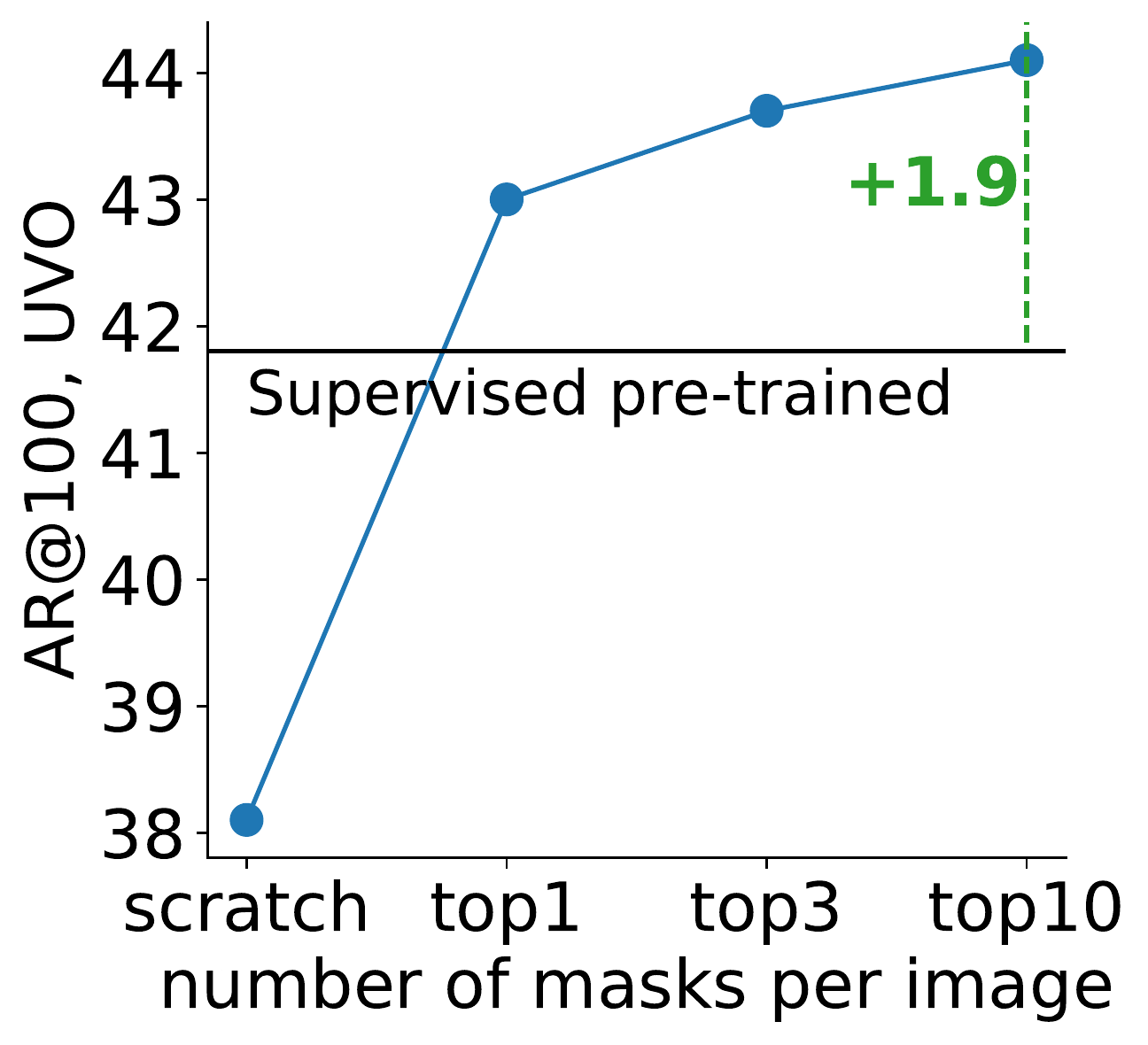}}
    \caption{UVO}
 \end{subfigure}
 \begin{subfigure}[b]{0.19\linewidth}
    {\includegraphics[width=\linewidth]{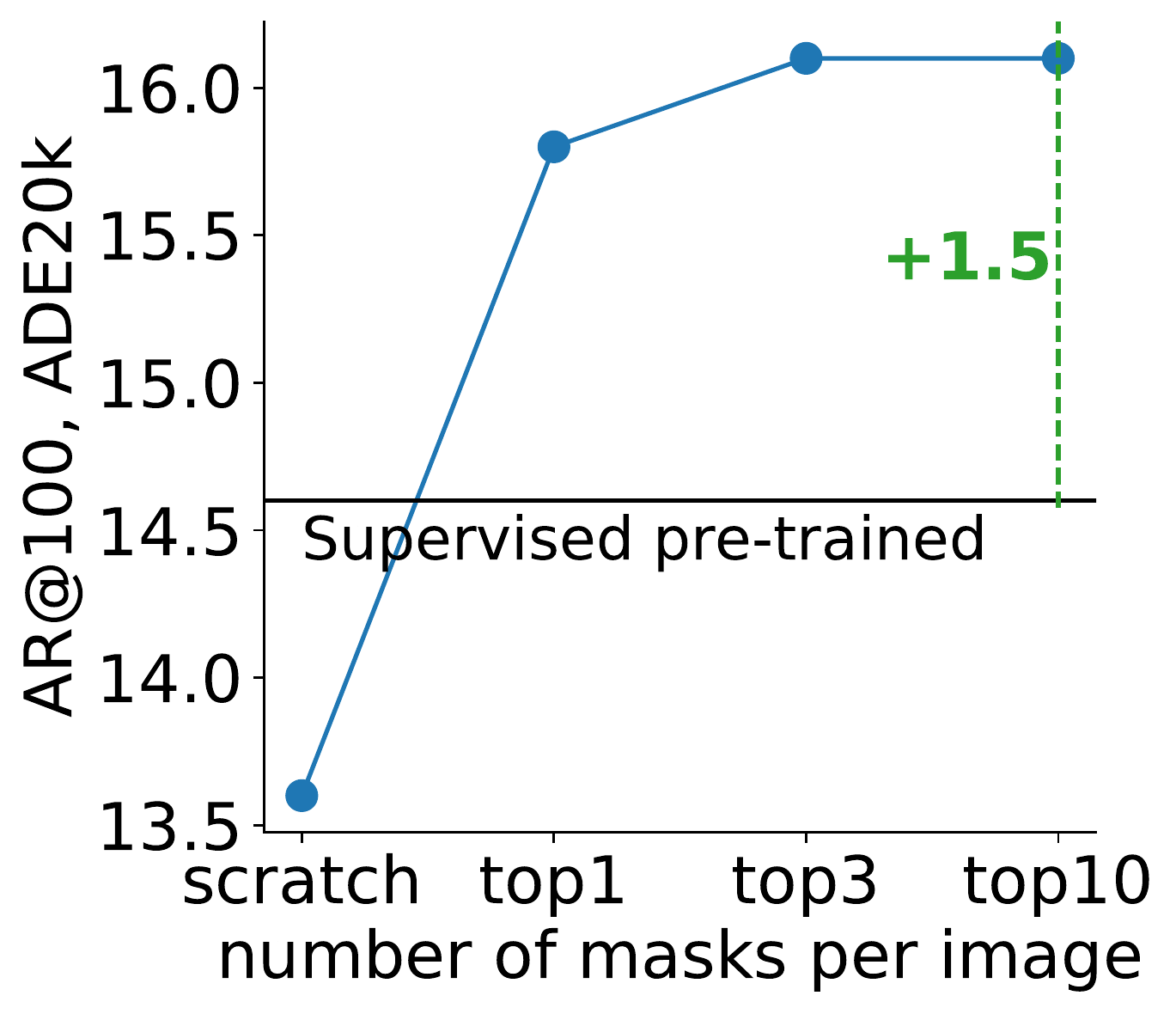}}
    \caption{ADE20K}
 \end{subfigure}
 \begin{subfigure}[b]{0.19\linewidth}
    {\includegraphics[width=\linewidth]{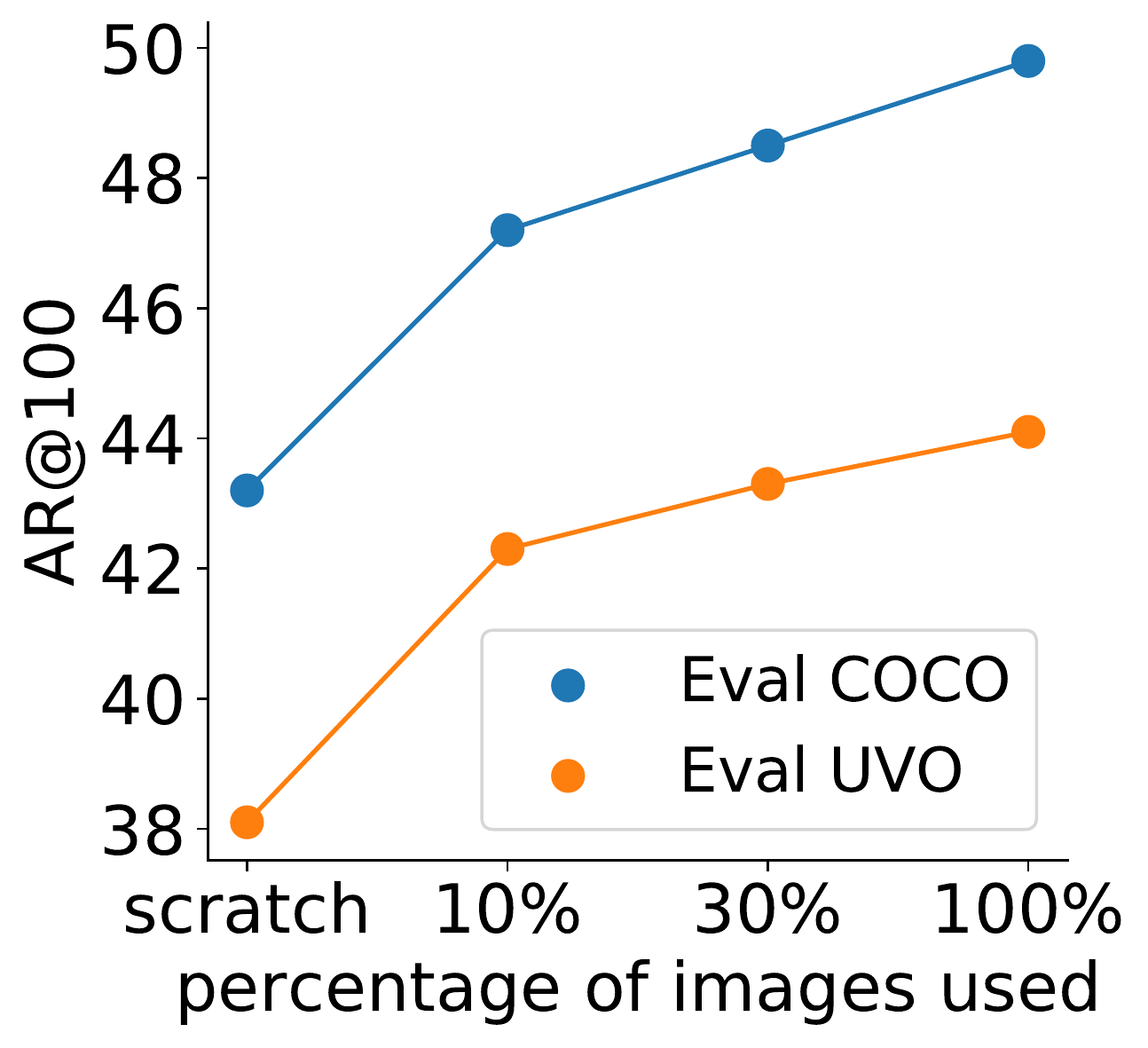}}
    \caption{Scaling pixels}
 \end{subfigure}
 \vspace{-8pt}
 \caption{{\bf \Our{} outperforms ImageNet supervised pre-training for open-world instance segmentation and demonstrates promising scaling behaviors}. We compare pseudo-GT mask pre-training by \Our{} with ImageNet label pre-training on closed-world (COCO, a), cross-category (non-COCO in LVIS, b) and open-world (UVO, ADE20K, c,d). Except closed-world setting, pseudo-GT masks provide stronger pre-training signals than ImageNet annotated labels (b-d). In addition, performance improves when more pseudo-GT masks selected per image (a-d) or more pixels (unlabeled images) (e) are used.
 }
 \label{fig:scale_imagenet}
 }
 \vspace{-8pt}
\end{figure*}

\begin{figure}
\centering
{\small 
 \begin{subfigure}[b]{0.32\linewidth}
    {\includegraphics[width=\linewidth]{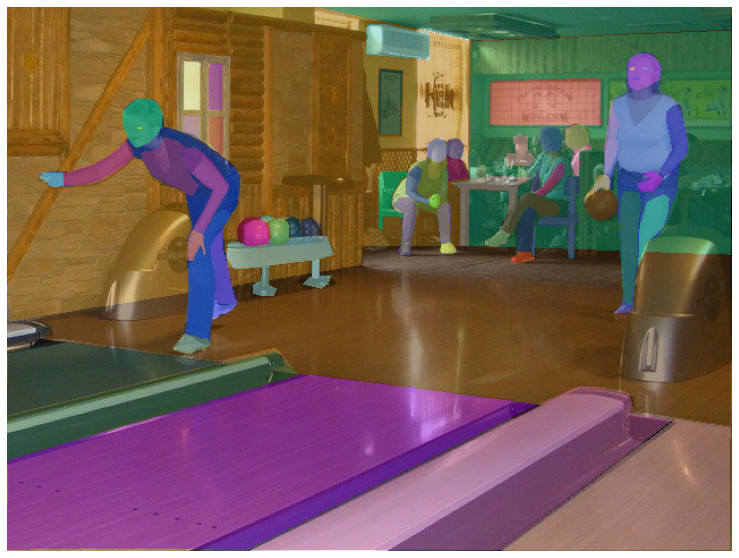}}
    \caption{GT (81)}
 \end{subfigure}
 \begin{subfigure}[b]{0.32\linewidth}
    {\includegraphics[width=\linewidth]{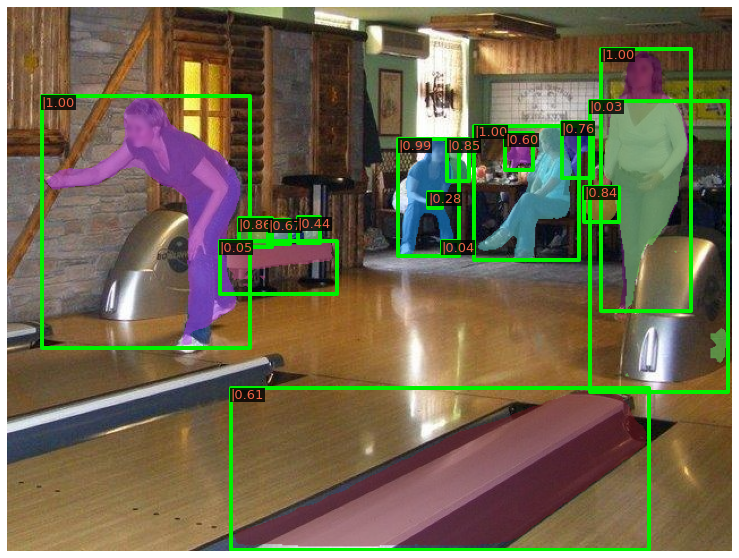}}
    \caption{Mask R-CNN (16)}
 \end{subfigure}
 \begin{subfigure}[b]{0.32\linewidth}
    {\includegraphics[width=\linewidth]{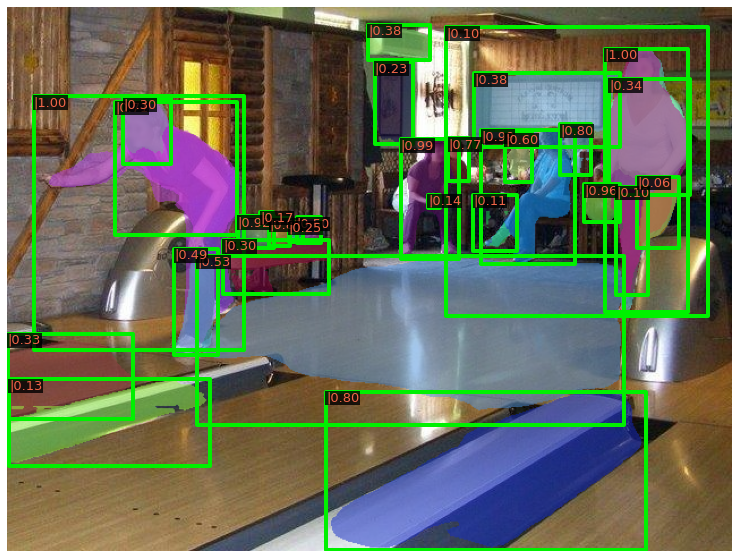}}
    \caption{\Our{} (30)}
 \end{subfigure}
 
 \begin{subfigure}[b]{0.32\linewidth}
    {\includegraphics[width=\linewidth]{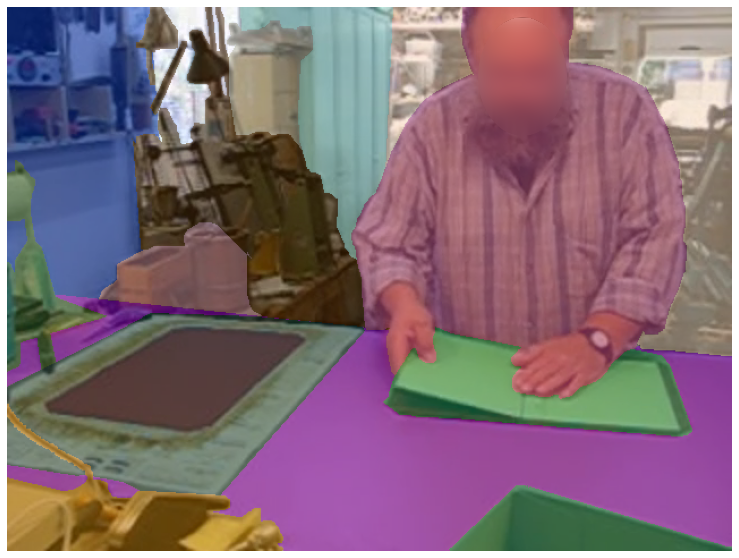}}
    \caption{GT (13)}
 \end{subfigure}
 \begin{subfigure}[b]{0.32\linewidth}
    {\includegraphics[width=\linewidth]{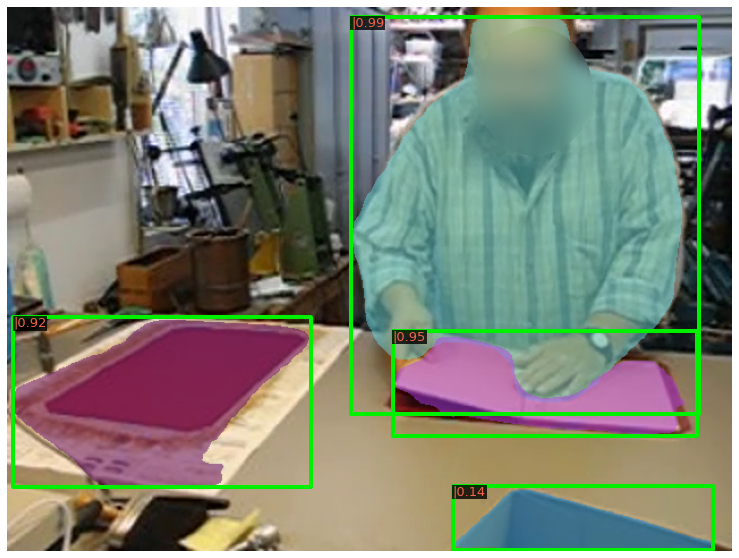}}
    \caption{Mask R-CNN (4)}
 \end{subfigure}
 \begin{subfigure}[b]{0.32\linewidth}
    {\includegraphics[width=\linewidth]{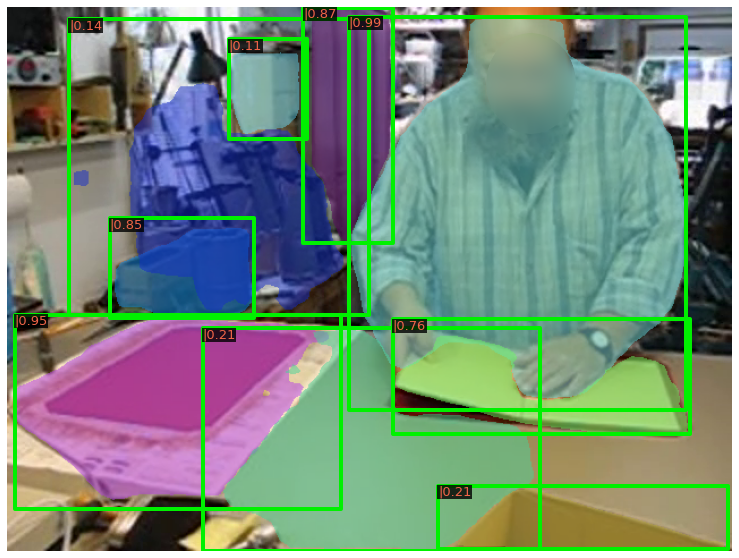}}
    \caption{\Our{} (9)}
 \end{subfigure}
 \vspace{-8pt}
 \caption{{\bf Visualization of predictions of \Our{} and Mask R-CNN on ADE20K and UVO}. \Our{} retrieves more instances correctly (numbers denoted in parentheses) and covers a more diverse set of object categories.
 }
 \label{fig:vis_ggn_pred}
 }
 \vspace{-8pt}
\end{figure}

Ablations in section~\ref{sec:cross_category} focus on cross-category generalization in a controlled version of open-world. A more practical question is: how well detectors can generalize across datasets in the wild? It is difficult to evaluate since common datasets, e.g., COCO and LVIS, are only \textit{partially-annotated}. Evaluating open-world segmentation on such datasets may fail to capture performance difference across methods due to punishing precision and not rewarding recall~\cite{Chavali2016ObjectProposalEP}. To handle this, we adopt ADE20K~\cite{8100027} and UVO~\cite{uvo} for evaluating generic proposals in the wild open-world. Specifically, we treat each segmentation mask in ADE20K or UVO as a ground-truth semantic entity and evaluate Average Recall (AR) and Average Precision (AP). This setup evaluates both in-taxonomy and out-taxonomy segments. While UVO contains only objects, ADE20K also includes stuff masks. We emphasize that this is truly {\bf in the wild test} as no fine-tuning is done on ADE20K or UVO.  

We compare \Our{} with Selective Search~\cite{SelectiveSearch} and Mask R-CNN baselines trained with the GT masks of all 80 COCO classes from COCO dataset (Table~\ref{tab:compare_wild}). \Our{} (enhanced by pseudo masks) significantly outperforms the baseline on both ADE20K and UVO dataset, both AR and AP. In addition, better ranking by combining $\mathcal{O}_{PA}$ and $\mathcal{O}_{OLN}$ further improves model performance. Qualitative results comparing Mask R-CNN and \Our{} on UVO and ADE20K are showed in Figure~\ref{fig:vis_ggn_pred} (more in supplementary).



\subsection{Pre-training on unlabeled images with \Our{}} \label{sec:scale_unlabeled}

Since PA-based bottom-up grouping can generate masks for any unlabeled image, we hypothesize that training with pseudo-GT masks from additional pixels may help open-world segmentation. Masks from PA generalizes well to new categories on new pixels, and thus benefit from pixel diversity. We study the effect of training \Our{} on pseudo-GT masks from unlabeled images from ImageNet~\cite{deng2009imagenet}. 

Specifically, we use PA trained on 80 COCO categories from COCO with random initialization. We generate pseudo-GT masks on ImageNet images and pre-train a randomly initialized Mask R-CNN on pseudo-GT masks (18 epochs). We then finetune the model on COCO annotated masks (80 categories) for standard 1x schedule. Similar to previous training from random initialization setup~\cite{He2019RethinkingIP}, we use GroupNorm~\cite{Wu2018GroupN} for long training with small batch size.

Results are summarized in Figure~\ref{fig:scale_imagenet}. When evaluated on COCO categories (same as training, closed-world), pre-training by pseudo-GT masks performs slightly worse than supervised label pre-training (-1.4\%). On open-world setup, however, pseudo-GT pre-training consistently outperforms supervised training. We note that different from closed-world~\cite{He2019RethinkingIP}, ImageNet supervised pre-training is a strong initialization for open-world (see supplementary). In addition, we observe two promising scaling behaviors of pseudo-GT pretraining: a. using more masks per image, despite some being noisy, improves performance; using more images/ pixels improves performance for both closed-world and open-world instance segmentation. We show similar results of Pre-training on images from OpenImages~\cite{OpenImages} in supplementary materials. Finetuning on both COCO annotated masks and PA generated Pseudo-masks on COCO images provides additional gain (last row in Table~\ref{tab:compare_wild}).

\section{Conclusion}
\label{sec:conclusion}

We have presented \Our{}, a novel approach for open-world instance segmentation which combines learned semantic boundaries with grouping to generate additional pseudo ground truth for instance-level training. \Ours{} significantly outperform baselines on various benchmarks. \Our{} is on par with state-of-the-art approaches, e.g., OLN~\cite{oln}, and when combined with OLN, \Our{} obtains an additional $6.8\%$, establishing new state-of-the-art results for open-world instance segmentation. Finally, we showed that \Our{} is robust when evaluated ``in the wild'' and benefits from training on additional unlabeled data.

\noindent \textbf{Acknowledgement.} We thank Ross Girshick for the discussion on baselines and grouping methods and Abhijit Ogale for the discussion about open-world setting.

{\small
\bibliographystyle{ieee_fullname}
\bibliography{egbib}

\begin{thebibliography}{10}\itemsep=-1pt

\bibitem{Achanta10slicsuperpixels}
Radhakrishna Achanta, Appu Shaji, Kevin Smith, Aurelien Lucchi, Pascal Fua, and
  Sabine Süsstrunk.
\newblock Slic superpixels, 2010.

\bibitem{AcunaCVPR19STEAL}
David Acuna, Amlan Kar, and Sanja Fidler.
\newblock Devil is in the edges: Learning semantic boundaries from noisy
  annotations.
\newblock In {\em CVPR}, 2019.

\bibitem{ucm}
Pablo Arbelaez.
\newblock Boundary extraction in natural images using ultrametric contour maps.
\newblock In {\em CVPR Workshops}, 2006.

\bibitem{ArbelaezMFM11}
Pablo Arbelaez, Michael Maire, Charless~C. Fowlkes, and Jitendra Malik.
\newblock Contour detection and hierarchical image segmentation.
\newblock {\em {IEEE} Trans. Pattern Anal. Mach. Intell.}, 33(5):898--916,
  2011.

\bibitem{ArbelaezPBMM14}
Pablo~Andr{\'{e}}s Arbel{\'{a}}ez, Jordi Pont{-}Tuset, Jonathan~T. Barron,
  Ferran Marqu{\'{e}}s, and Jitendra Malik.
\newblock Multiscale combinatorial grouping.
\newblock In {\em CVPR}, 2014.

\bibitem{bendale2015towards}
Abhijit Bendale and Terrance Boult.
\newblock Towards open world recognition.
\newblock In {\em Proceedings of the IEEE conference on computer vision and
  pattern recognition}, pages 1893--1902, 2015.

\bibitem{SEEDS3D}
M.~V.~D. {Bergh}, G. {Roig}, X. {Boix}, S. {Manen}, and L.~V. {Gool}.
\newblock Online video seeds for temporal window objectness.
\newblock In {\em 2013 IEEE International Conference on Computer Vision}, pages
  377--384, 2013.

\bibitem{BoykovVZ01}
Yuri Boykov, Olga Veksler, and Ramin Zabih.
\newblock Fast approximate energy minimization via graph cuts.
\newblock {\em {IEEE} Trans. Pattern Anal. Mach. Intell.}, 23(11):1222--1239,
  2001.

\bibitem{Chavali2016ObjectProposalEP}
Neelima Chavali, Harsh Agrawal, Aroma Mahendru, and Dhruv Batra.
\newblock Object-proposal evaluation protocol is ‘gameable’.
\newblock {\em 2016 IEEE Conference on Computer Vision and Pattern Recognition
  (CVPR)}, pages 835--844, 2016.

\bibitem{mmdetection}
Kai Chen, Jiaqi Wang, Jiangmiao Pang, Yuhang Cao, Yu Xiong, Xiaoxiao Li,
  Shuyang Sun, Wansen Feng, Ziwei Liu, Jiarui Xu, Zheng Zhang, Dazhi Cheng,
  Chenchen Zhu, Tianheng Cheng, Qijie Zhao, Buyu Li, Xin Lu, Rui Zhu, Yue Wu,
  Jifeng Dai, Jingdong Wang, Jianping Shi, Wanli Ouyang, Chen~Change Loy, and
  Dahua Lin.
\newblock {MMDetection}: Open mmlab detection toolbox and benchmark.
\newblock {\em arXiv preprint arXiv:1906.07155}, 2019.

\bibitem{chen2020simple}
Ting Chen, Simon Kornblith, Mohammad Norouzi, and Geoffrey Hinton.
\newblock A simple framework for contrastive learning of visual
  representations.
\newblock {\em arXiv preprint arXiv:2002.05709}, 2020.

\bibitem{chen2020mocov2}
Xinlei Chen, Haoqi Fan, Ross Girshick, and Kaiming He.
\newblock Improved baselines with momentum contrastive learning.
\newblock {\em arXiv preprint arXiv:2003.04297}, 2020.

\bibitem{deng2009imagenet}
Jia Deng, Wei Dong, Richard Socher, Li-Jia Li, Kai Li, and Li Fei-Fei.
\newblock Imagenet: A large-scale hierarchical image database.
\newblock In {\em CVPR}, pages 248--255, 2009.

\bibitem{EndresH10}
Ian Endres and Derek Hoiem.
\newblock Category independent object proposals.
\newblock In {\em ECCV}, 2010.

\bibitem{PascalVOC}
M. Everingham, S.~M.~A. Eslami, L. Van~Gool, C.~K.~I. Williams, J. Winn, and A.
  Zisserman.
\newblock The pascal visual object classes challenge: A retrospective.
\newblock {\em International Journal of Computer Vision}, 111(1):98--136, Jan.
  2015.

\bibitem{gb}
Pedro~F. Felzenszwalb and Daniel~P. Huttenlocher.
\newblock Efficient graph-based image segmentation.
\newblock {\em IJCV}, 59(2):167--181, 2004.

\bibitem{FowlkesMM03}
Charless~C. Fowlkes, David~R. Martin, and Jitendra Malik.
\newblock Learning affinity functions for image segmentation: Combining
  patch-based and gradient-based approaches.
\newblock In {\em CVPR}, 2003.

\bibitem{Gao2019SSAPSI}
Naiyu Gao, Yanhu Shan, Yupei Wang, Xin Zhao, Yinan Yu, Ming Yang, and Kaiqi
  Huang.
\newblock Ssap: Single-shot instance segmentation with affinity pyramid.
\newblock {\em 2019 IEEE/CVF International Conference on Computer Vision
  (ICCV)}, pages 642--651, 2019.

\bibitem{GBH}
M. {Grundmann}, V. {Kwatra}, M. {Han}, and I. {Essa}.
\newblock Efficient hierarchical graph-based video segmentation.
\newblock In {\em 2010 IEEE Computer Society Conference on Computer Vision and
  Pattern Recognition}, pages 2141--2148, 2010.

\bibitem{8954457}
A. {Gupta}, P. {Dollár}, and R. {Girshick}.
\newblock Lvis: A dataset for large vocabulary instance segmentation.
\newblock In {\em 2019 IEEE/CVF Conference on Computer Vision and Pattern
  Recognition (CVPR)}, pages 5351--5359, 2019.

\bibitem{He2020MomentumCF}
Kaiming He, Haoqi Fan, Yuxin Wu, Saining Xie, and Ross~B. Girshick.
\newblock Momentum contrast for unsupervised visual representation learning.
\newblock {\em 2020 IEEE/CVF Conference on Computer Vision and Pattern
  Recognition (CVPR)}, pages 9726--9735, 2020.

\bibitem{He2019RethinkingIP}
Kaiming He, Ross~B. Girshick, and Piotr Doll{\'a}r.
\newblock Rethinking imagenet pre-training.
\newblock {\em 2019 IEEE/CVF International Conference on Computer Vision
  (ICCV)}, pages 4917--4926, 2019.

\bibitem{8237584}
K. {He}, G. {Gkioxari}, P. {Dollár}, and R. {Girshick}.
\newblock Mask r-cnn.
\newblock In {\em ICCV}, pages 2980--2988, 2017.

\bibitem{HeZC04}
Xuming He, Richard~S. Zemel, and Miguel~{\'{A}}.
  Carreira{-}Perpi{\~{n}}{\'{a}}n.
\newblock Multiscale conditional random fields for image labeling.
\newblock In {\em CVPR}, 2004.

\bibitem{868688}
{Jianbo Shi} and J. {Malik}.
\newblock Normalized cuts and image segmentation.
\newblock {\em IEEE Transactions on Pattern Analysis and Machine Intelligence},
  22(8):888--905, 2000.

\bibitem{joseph2021open}
K~J Joseph, Salman Khan, Fahad~Shahbaz Khan, and Vineeth~N Balasubramanian.
\newblock Towards open world object detection.
\newblock In {\em CVPR}, 2021.

\bibitem{kinetics}
Will Kay, Joao Carreira, Karen Simonyan, Brian Zhang, Chloe Hillier, Sudheendra
  Vijayanarasimhan, Fabio Viola, Tim Green, Trevor Back, Paul Natsev, Mustafa
  Suleyman, and Andrew Zisserman.
\newblock The kinetics human action video dataset.
\newblock {\em CoRR}, abs/1705.06950, 2017.

\bibitem{oln}
Dahun Kim, Tsung{-}Yi Lin, Anelia Angelova, In~So Kweon, and Weicheng Kuo.
\newblock Learning open-world object proposals without learning to classify.
\newblock {\em CoRR}, abs/2108.06753, 2021.

\bibitem{KimLL13}
Tae~Hoon Kim, Kyoung~Mu Lee, and Sang~Uk Lee.
\newblock Learning full pairwise affinities for spectral segmentation.
\newblock {\em {IEEE} Trans. Pattern Anal. Mach. Intell.}, 35(7):1690--1703,
  2013.

\bibitem{KirillovHGRD19}
Alexander Kirillov, Kaiming He, Ross~B. Girshick, Carsten Rother, and Piotr
  Doll{\'{a}}r.
\newblock Panoptic segmentation.
\newblock In {\em CVPR}, 2019.

\bibitem{Kong_2021_ICCV}
Shu Kong and Deva Ramanan.
\newblock Opengan: Open-set recognition via open data generation.
\newblock In {\em Proceedings of the IEEE/CVF International Conference on
  Computer Vision (ICCV)}, pages 813--822, October 2021.

\bibitem{OpenImages}
Alina Kuznetsova, Hassan Rom, Neil Alldrin, Jasper Uijlings, Ivan Krasin, Jordi
  Pont-Tuset, Shahab Kamali, Stefan Popov, Matteo Malloci, Alexander
  Kolesnikov, Tom Duerig, and Vittorio Ferrari.
\newblock The open images dataset v4: Unified image classification, object
  detection, and visual relationship detection at scale.
\newblock {\em IJCV}, 2020.

\bibitem{li2019learning}
Zhengqi Li, Tali Dekel, Forrester Cole, Richard Tucker, Noah Snavely, Ce Liu,
  and William~T Freeman.
\newblock Learning the depths of moving people by watching frozen people.
\newblock In {\em Proc. Computer Vision and Pattern Recognition (CVPR)}, 2019.

\bibitem{Lin2014MicrosoftCC}
Tsung-Yi Lin, M. Maire, Serge~J. Belongie, James Hays, P. Perona, D. Ramanan,
  Piotr Doll{\'a}r, and C.~L. Zitnick.
\newblock Microsoft coco: Common objects in context.
\newblock In {\em ECCV}, 2014.

\bibitem{LiuMGZ0K17}
Sifei Liu, Shalini~De Mello, Jinwei Gu, Guangyu Zhong, Ming{-}Hsuan Yang, and
  Jan Kautz.
\newblock Learning affinity via spatial propagation networks.
\newblock In {\em Advances in Neural Information Processing Systems 30: Annual
  Conference on Neural Information Processing Systems 2017, December 4-9, 2017,
  Long Beach, CA, {USA}}, 2017.

\bibitem{Liu2018SemanticED}
Yun Liu, Ming-Ming Cheng, Jiawang Bian, Le Zhang, Peng-Tao Jiang, and Yang Cao.
\newblock Semantic edge detection with diverse deep supervision.
\newblock {\em ArXiv}, abs/1804.02864, 2018.

\bibitem{Liu2018AffinityDA}
Yiding Liu, Si~Cheng. Yang, Bin Li, Wen gang Zhou, Jizheng Xu, Houqiang Li, and
  Yan Lu.
\newblock Affinity derivation and graph merge for instance segmentation.
\newblock In {\em ECCV}, 2018.

\bibitem{abs-2104-11221}
Yang Liu, Idil~Esen Zulfikar, Jonathon Luiten, Achal Dave, Aljosa Osep, Deva
  Ramanan, Bastian Leibe, and Laura Leal{-}Taix{\'{e}}.
\newblock Opening up open-world tracking.
\newblock {\em CoRR}, abs/2104.11221, 2021.

\bibitem{swin_transformer}
Ze Liu, Yutong Lin, Yue Cao, Han Hu, Yixuan Wei, Zheng Zhang, Stephen Lin, and
  Baining Guo.
\newblock Swin transformer: Hierarchical vision transformer using shifted
  windows.
\newblock In {\em ICCV}, 2021.

\bibitem{liu2019large}
Ziwei Liu, Zhongqi Miao, Xiaohang Zhan, Jiayun Wang, Boqing Gong, and Stella~X
  Yu.
\newblock Large-scale long-tailed recognition in an open world.
\newblock In {\em Proceedings of the IEEE/CVF Conference on Computer Vision and
  Pattern Recognition}, pages 2537--2546, 2019.

\bibitem{fcn}
Jonathan Long, Evan Shelhamer, and Trevor Darrell.
\newblock Fully convolutional networks for semantic segmentation.
\newblock In {\em CVPR}, 2015.

\bibitem{Maire_2016_CVPR}
Michael Maire, Takuya Narihira, and Stella~X. Yu.
\newblock Affinity cnn: Learning pixel-centric pairwise relations for
  figure/ground embedding.
\newblock In {\em Proceedings of the IEEE Conference on Computer Vision and
  Pattern Recognition (CVPR)}, June 2016.

\bibitem{COB}
Kevis-Kokitsi Maninis, Jordi Pont-Tuset, Pablo Arbel\'{a}ez, and Luc~Van Gool.
\newblock Convolutional oriented boundaries.
\newblock In {\em European Conference on Computer Vision (ECCV)}, 2016.

\bibitem{Pathak2016ContextEF}
Deepak Pathak, Philipp Kr{\"a}henb{\"u}hl, Jeff Donahue, Trevor Darrell, and
  Alexei~A. Efros.
\newblock Context encoders: Feature learning by inpainting.
\newblock {\em 2016 IEEE Conference on Computer Vision and Pattern Recognition
  (CVPR)}, pages 2536--2544, 2016.

\bibitem{DeepMask15}
Pedro~O. Pinheiro, Ronan Collobert, and Piotr Doll\'{a}r.
\newblock Learning to segment object candidates.
\newblock In {\em Proceedings of the 28th International Conference on Neural
  Information Processing Systems - Volume 2}, NIPS'15, page 1990–1998,
  Cambridge, MA, USA, 2015. MIT Press.

\bibitem{PIZER1987355}
Stephen~M. Pizer, E.~Philip Amburn, John~D. Austin, Robert Cromartie, Ari
  Geselowitz, Trey Greer, Bart {ter Haar Romeny}, John~B. Zimmerman, and Karel
  Zuiderveld.
\newblock Adaptive histogram equalization and its variations.
\newblock {\em Computer Vision, Graphics, and Image Processing},
  39(3):355--368, 1987.

\bibitem{7780460}
J. {Redmon}, S. {Divvala}, R. {Girshick}, and A. {Farhadi}.
\newblock You only look once: Unified, real-time object detection.
\newblock In {\em 2016 IEEE Conference on Computer Vision and Pattern
  Recognition (CVPR)}, pages 779--788, 2016.

\bibitem{FasterRCNN}
Shaoqing Ren, Kaiming He, Ross Girshick, and Jian Sun.
\newblock Faster r-cnn: Towards real-time object detection with region proposal
  networks.
\newblock In C. Cortes, N. Lawrence, D. Lee, M. Sugiyama, and R. Garnett,
  editors, {\em Advances in Neural Information Processing Systems}, volume~28,
  pages 91--99. Curran Associates, Inc., 2015.

\bibitem{10.5555/1864519.1864542}
Ellen Riloff.
\newblock Automatically generating extraction patterns from untagged text.
\newblock In {\em Proceedings of the Thirteenth National Conference on
  Artificial Intelligence - Volume 2}, AAAI'96, page 1044–1049. AAAI Press,
  1996.

\bibitem{1053799}
H. Scudder.
\newblock Probability of error of some adaptive pattern-recognition machines.
\newblock {\em IEEE Transactions on Information Theory}, 11(3):363--371, 1965.

\bibitem{ShiM00}
Jianbo Shi and Jitendra Malik.
\newblock Normalized cuts and image segmentation.
\newblock {\em {IEEE} Trans. Pattern Anal. Mach. Intell.}, 22(8):888--905,
  2000.

\bibitem{sohn2020detection}
Kihyuk Sohn, Zizhao Zhang, Chun-Liang Li, Han Zhang, Chen-Yu Lee, and Tomas
  Pfister.
\newblock A simple semi-supervised learning framework for object detection.
\newblock In {\em arXiv:2005.04757}, 2020.

\bibitem{soria2020dexined}
Xavier Soria, Edgar Riba, and Angel Sappa.
\newblock Dense extreme inception network: Towards a robust cnn model for edge
  detection.
\newblock In {\em The IEEE Winter Conference on Applications of Computer Vision
  (WACV '20)}, 2020.

\bibitem{TuragaBHDS09}
Srinivas~C. Turaga, Kevin~L. Briggman, Moritz Helmstaedter, Winfried Denk, and
  H.~Sebastian Seung.
\newblock Maximin affinity learning of image segmentation.
\newblock In {\em Neural Information Processing Systems}, 2009.

\bibitem{SelectiveSearch}
K.~E.~A. {van de Sande}, J.~R.~R. {Uijlings}, T. {Gevers}, and A.~W.~M.
  {Smeulders}.
\newblock Segmentation as selective search for object recognition.
\newblock In {\em ICCV}, pages 1879--1886, 2011.

\bibitem{MaX-DeepLab}
Huiyu Wang, Yukun Zhu, Hartwig Adam, Alan~L. Yuille, and Liang{-}Chieh Chen.
\newblock Max-deeplab: End-to-end panoptic segmentation with mask transformers.
\newblock In {\em {IEEE} Conference on Computer Vision and Pattern Recognition,
  {CVPR} 2021, virtual, June 19-25, 2021}, 2021.

\bibitem{Wang2020WhatLT}
Rui Wang, Dhruv~Kumar Mahajan, and Vignesh Ramanathan.
\newblock What leads to generalization of object proposals?
\newblock In {\em ECCV Workshops}, 2020.

\bibitem{uvo}
Weiyao Wang, Matt Feiszli, Heng Wang, and Du Tran.
\newblock Unidentified video objects: A benchmark for dense, open-world
  segmentation.
\newblock In {\em ICCV}, 2021.

\bibitem{Wu2018GroupN}
Yuxin Wu and Kaiming He.
\newblock Group normalization.
\newblock In {\em ECCV}, 2018.

\bibitem{UPerNet}
Tete Xiao, Yingcheng Liu, Bolei Zhou, Yuning Jiang, and Jian Sun.
\newblock Unified perceptual parsing for scene understanding.
\newblock In {\em ECCV}, 2018.

\bibitem{7410521}
Saining Xie and Zhuowen Tu.
\newblock Holistically-nested edge detection.
\newblock In {\em 2015 IEEE International Conference on Computer Vision
  (ICCV)}, pages 1395--1403, 2015.

\bibitem{StreamGBH}
Chenliang Xu, Caiming Xiong, and Jason~J. Corso.
\newblock Streaming hierarchical video segmentation.
\newblock In Andrew Fitzgibbon, Svetlana Lazebnik, Pietro Perona, Yoichi Sato,
  and Cordelia Schmid, editors, {\em Computer Vision -- ECCV 2012}, pages
  626--639, Berlin, Heidelberg, 2012. Springer Berlin Heidelberg.

\bibitem{10.3115/981658.981684}
David Yarowsky.
\newblock Unsupervised word sense disambiguation rivaling supervised methods.
\newblock In {\em Proceedings of the 33rd Annual Meeting on Association for
  Computational Linguistics}, ACL '95, page 189–196, USA, 1995. Association
  for Computational Linguistics.

\bibitem{ZhangM20}
Xiao Zhang and Michael Maire.
\newblock Self-supervised visual representation learning from hierarchical
  grouping.
\newblock In {\em NeurIPS}, 2020.

\bibitem{8100027}
B. {Zhou}, H. {Zhao}, X. {Puig}, S. {Fidler}, A. {Barriuso}, and A. {Torralba}.
\newblock Scene parsing through ade20k dataset.
\newblock In {\em 2017 IEEE Conference on Computer Vision and Pattern
  Recognition (CVPR)}, pages 5122--5130, 2017.

\bibitem{Zoph2020LearningDA}
Barret Zoph, Ekin~Dogus Cubuk, Golnaz Ghiasi, Tsung-Yi Lin, Jonathon Shlens,
  and Quoc~V. Le.
\newblock Learning data augmentation strategies for object detection.
\newblock In {\em ECCV}, 2020.

\bibitem{Zoph2020RethinkingST}
Barret Zoph, Golnaz Ghiasi, Tsung-Yi Lin, Yin Cui, Hanxiao Liu, Ekin~Dogus
  Cubuk, and Quoc~V. Le.
\newblock Rethinking pre-training and self-training.
\newblock {\em ArXiv}, abs/2006.06882, 2020.

\end{thebibliography}
}

\clearpage

\pagebreak

\appendix

\noindent In this supplementary material, we include: 
\begin{enumerate}
    \item Experiments on OpenImages~\cite{OpenImages} demonstrating that scaling number of unlabeled training images further improves \Our{} (section~\ref{sec:scaling}).
    \item Effects of ImageNet~\cite{deng2009imagenet} pre-training on open-world instance segmentation (section~\ref{sec:imagenet_pretrain})
    \item Additional qualitative results of open-world segmentation in the wild on ADE20K and UVO (section~\ref{sec:qual}).
    \item A proof of concept experiment of using \Our{} for closed-world class-aware instance segmentation (section~\ref{sec:two_tower}).
    \item Our discussion on the limitations and future directions of the proposed method (section~\ref{sec:limit}).
    \item Ablations on data augmentation techniques for training PA (section~\ref{sec:PA_data_aug})
\end{enumerate}

\section{Improve \Our{} by scaling unlabeled pixels}
\label{sec:scaling}

In section 5 of main paper, we showed how our proposed method generates pseudo-GT masks on unlabeled images, and how \Our{} benefited from training on unlabeled images (Table 7). Here, we further show how \Our{} can be further improved by scaling the number of unlabeled training images.

\begin{figure}
\centering
{\small 
 \begin{subfigure}[b]{0.49\linewidth}
    {\includegraphics[width=\linewidth]{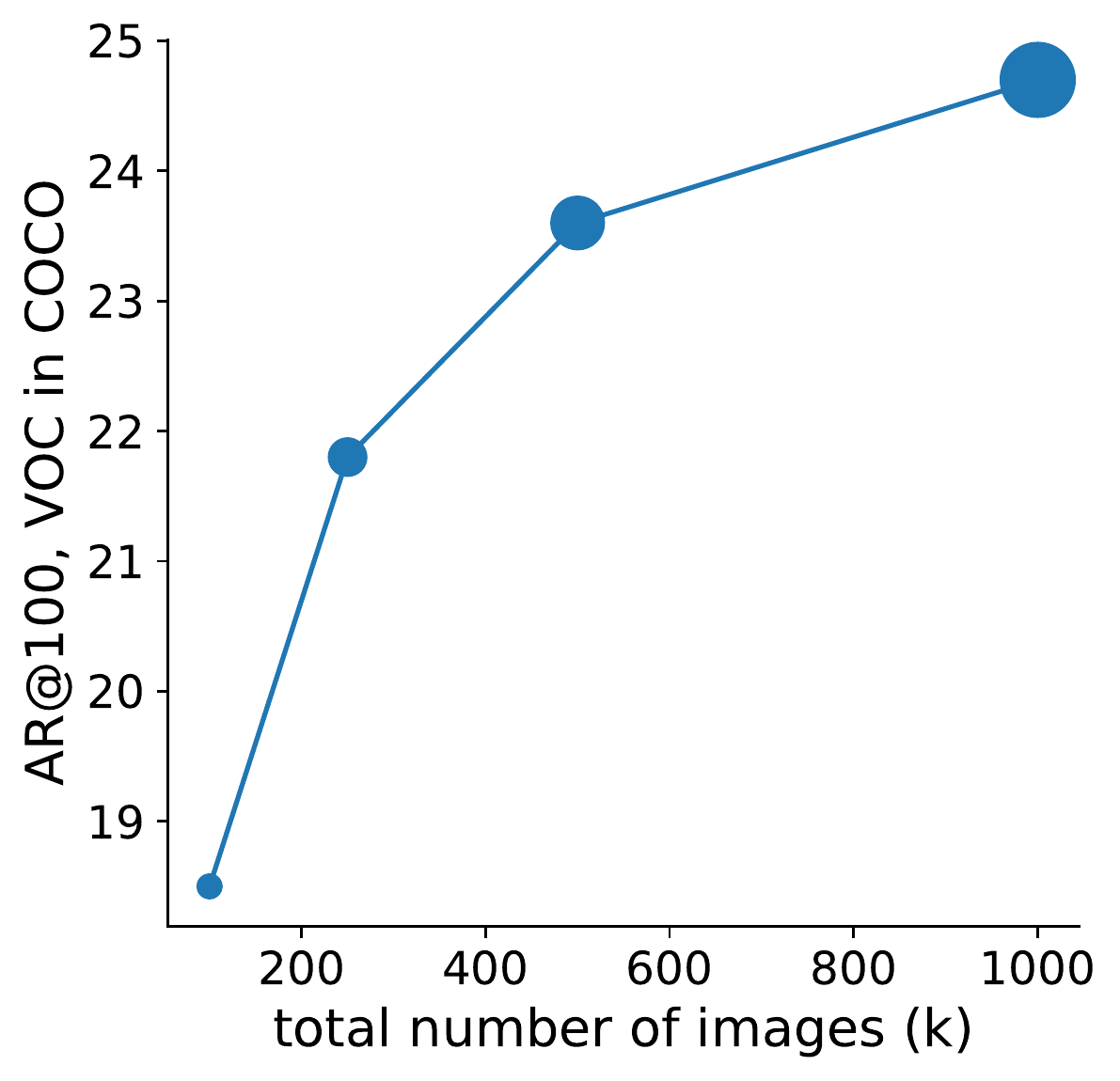}}
    \caption{}
 \end{subfigure}
 \begin{subfigure}[b]{0.49\linewidth}
    {\includegraphics[width=\linewidth]{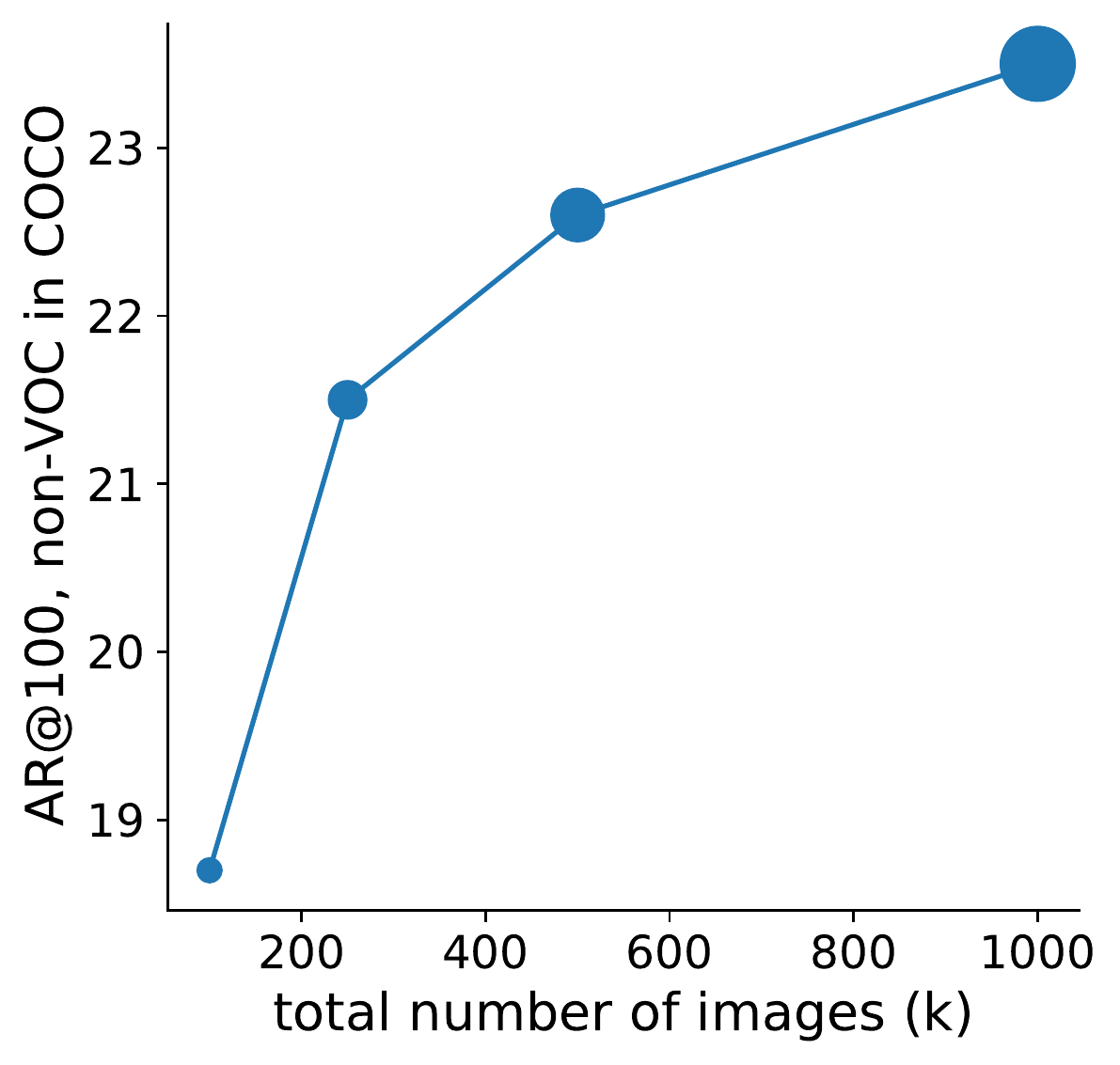}}
    \caption{}
 \end{subfigure}
 \begin{subfigure}[b]{0.49\linewidth}
    {\includegraphics[width=\linewidth]{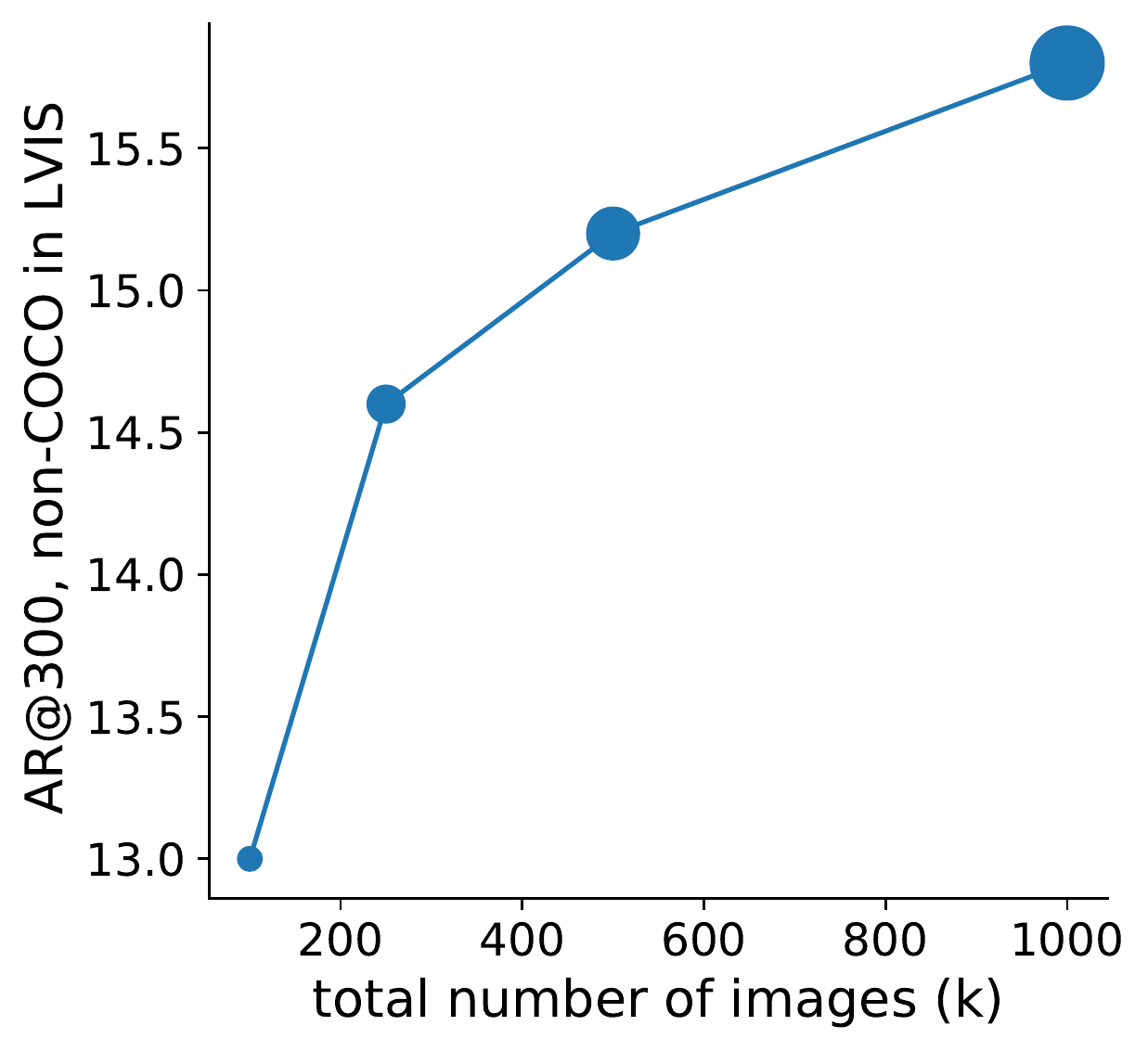}}
    \caption{}
 \end{subfigure}
 \begin{subfigure}[b]{0.49\linewidth}
    {\includegraphics[width=\linewidth]{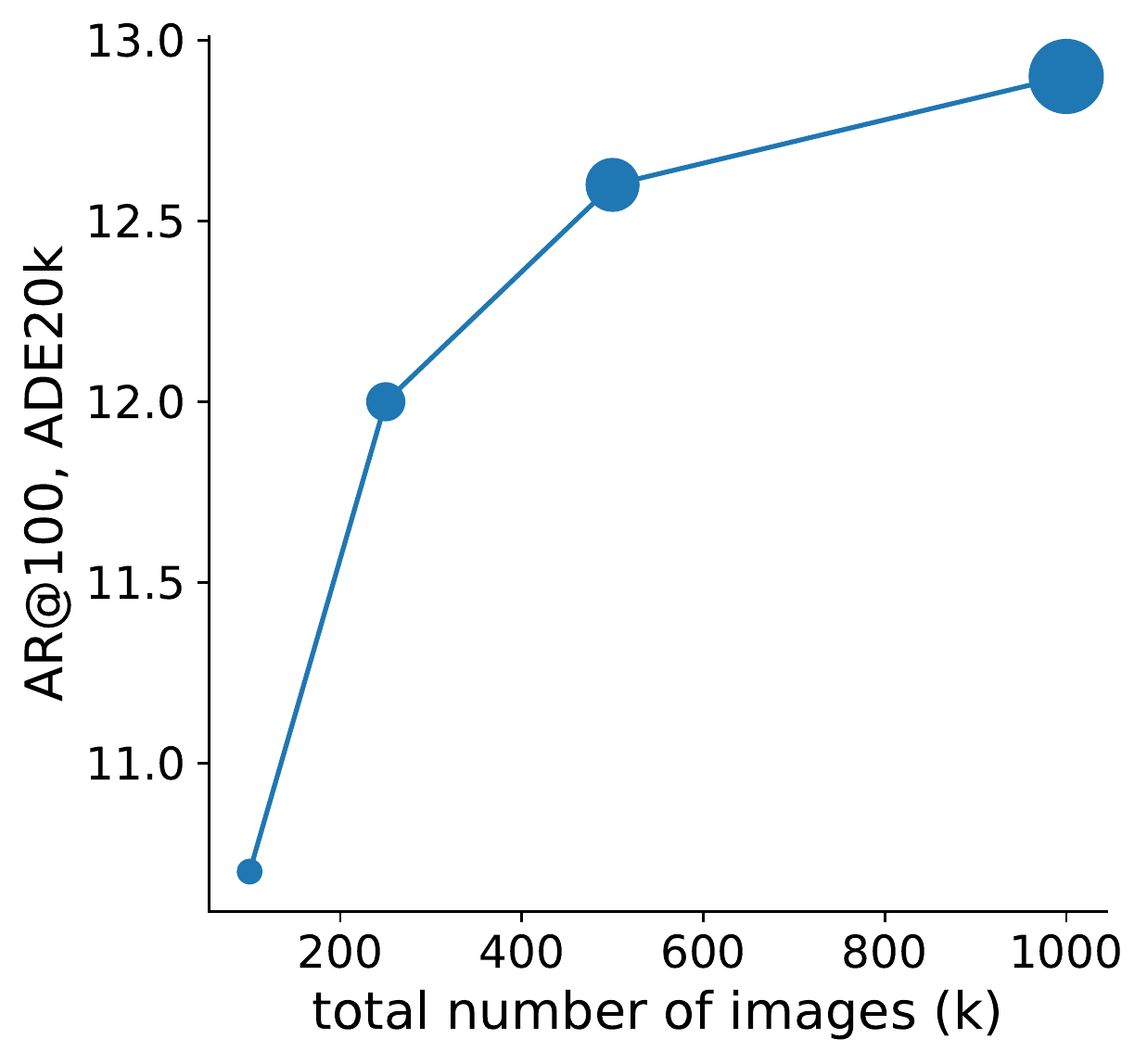}}
    \caption{}
 \end{subfigure}
 \vspace{-8pt}
 \caption{{\bf The effect of scaling the number of images in training \Our{}.} We increase the size of subset of OpenImages~\cite{OpenImages} to 100k, 250k, 500k and 1M and train \Ours{} with pseudo-masks generated by pairwise affinities trained on VOC masks. In all setups, scaling images keeps improving model performance. 
 }
 \label{fig:scale_pixels}
 }
\end{figure}

We increase the size of unlabeled images (e.g., 100k, 250k, 500k, 1M) sampled from OpenImagesV4~\cite{OpenImages} and take top-3 scoring pseudo-masks per image and use them as pseudo-GT masks for training. As shown in Figure~\ref{fig:scale_pixels}, increasing the number of unlabelled training images continuously improves model performances in various setups. This further demonstrates the potential of \Our{} in both open-world (non-VOC, non-COCO~\cite{Lin2014MicrosoftCC}, ADE20k~\cite{8100027}) and closed-world (VOC) instance segmentation.

\section{ImageNet pre-training for open-world instance segmentation} \label{sec:imagenet_pretrain}

In closed-world setup, ImageNet label pre-training offers limited values~\cite{He2019RethinkingIP}: when training from scratch at 6x standard schedule, detectors perform on-par with 1x schedule finetuning from ImageNet label pre-training. This questioned if ImageNet label pre-training is a strong baseline to compare to (as we did in section 5.5). We argue that it is indeed a strong baseline, and that ImageNet label pre-training outperforms 6x schedule training from scratch (Table~\ref{tab:imagenet_pretrain}). This validates the value of ImageNet label pre-training for open-world instance segmentation, making it a proper baseline to compare with.

\begin{table}
\captionsetup{font=footnotesize}
  \centering
  {\small
  \begin{tabular}{|c|c|c|c|}
    \hline
    Training strategy & LVIS & UVO & ADE20K  \\
    \hline
    6x schedule from scratch & 13.1 & 41.5 & 13.7 \\
    ImageNet pre-training & \textbf{14.7} & \textbf{41.8} & \textbf{14.6} \\
    \hline
  \end{tabular}}
  \caption{\textbf{Different from common wisdom in closed-world instance segmentation, ImageNet pre-training outperforms long training schedule from random initialization in open-world}. We verify this with Mask R-CNN trained/ finetuned on COCO and evaluate on Non-COCO categories in LVIS~\cite{8954457}, UVO~\cite{uvo} and ADE20K~\cite{8100027}}
  \label{tab:imagenet_pretrain}
\end{table}

\section{Qualitative results in the wild}
\label{sec:qual}

\begin{figure*}
\centering
{\small 
 
 \begin{subfigure}[b]{0.24\linewidth}
    {\includegraphics[width=\linewidth]{latex/images/ade_val_vis/idx10/orig.png}}
    \caption{Original Image}
 \end{subfigure}
 \begin{subfigure}[b]{0.24\linewidth}
    {\includegraphics[width=\linewidth]{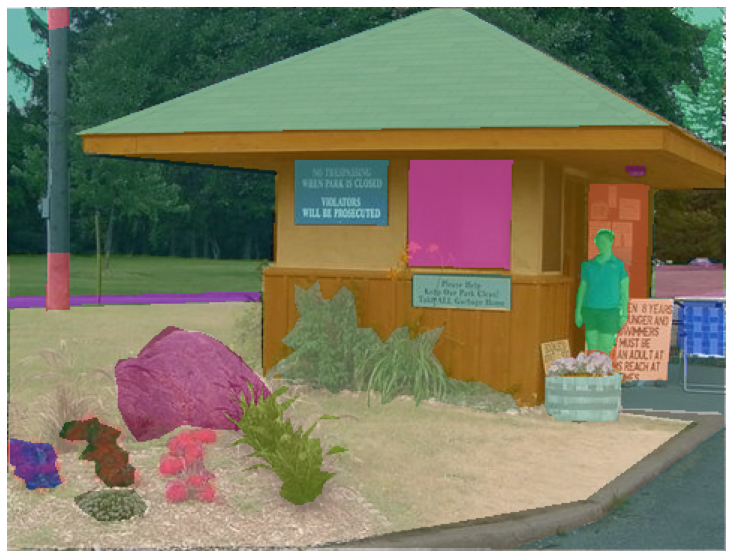}}
    \caption{GT (31)}
 \end{subfigure}
 \begin{subfigure}[b]{0.24\linewidth}
    {\includegraphics[width=\linewidth]{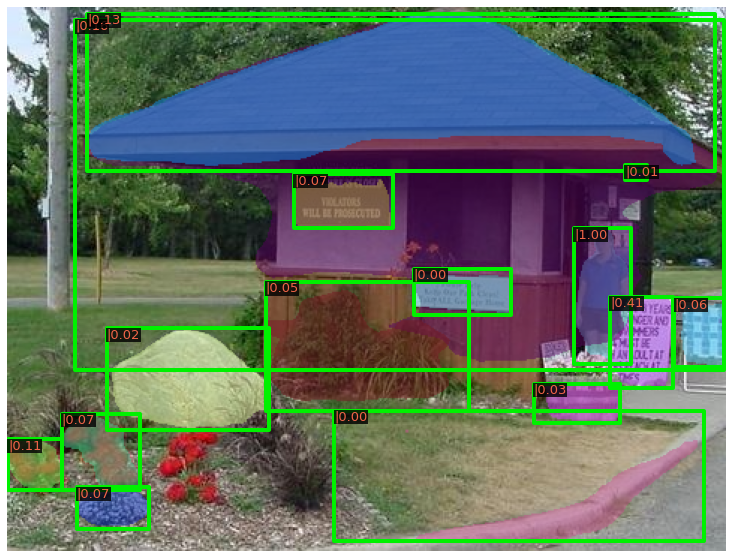}}
    \caption{Baseline (15)}
 \end{subfigure}
 \begin{subfigure}[b]{0.24\linewidth}
    {\includegraphics[width=\linewidth]{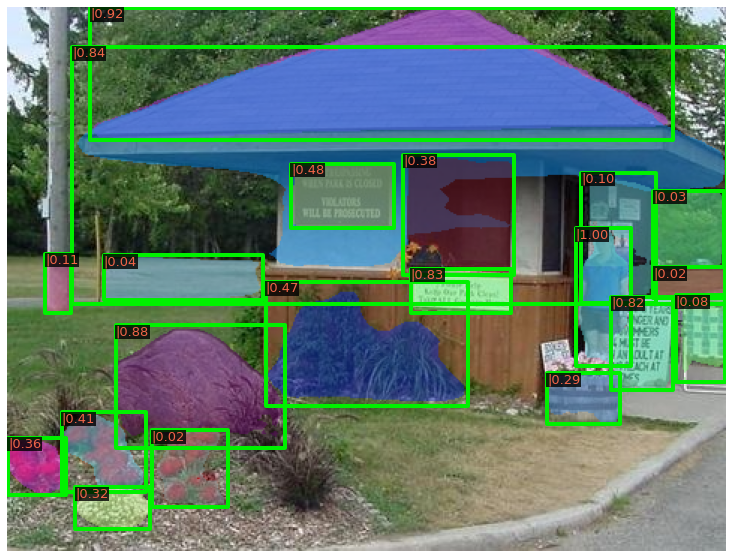}}
    \caption{\Our{} (20)}
 \end{subfigure}
 
 \begin{subfigure}[b]{0.24\linewidth}
    {\includegraphics[width=\linewidth]{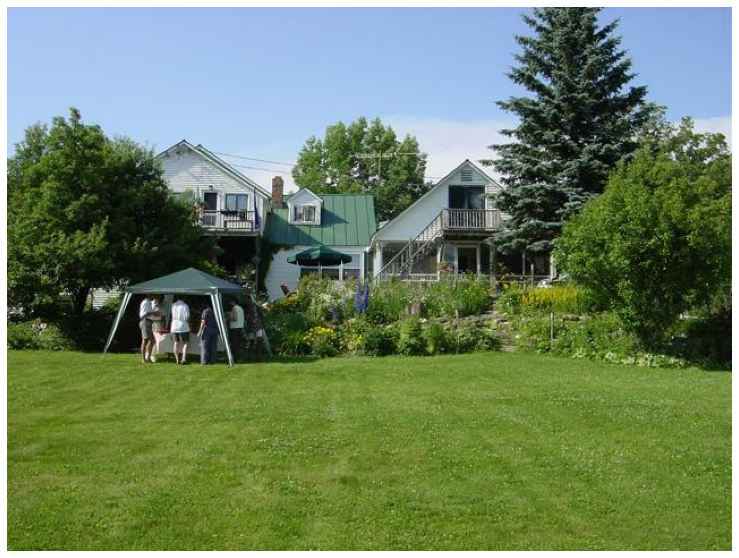}}
    \caption{Original Image}
 \end{subfigure}
 \begin{subfigure}[b]{0.24\linewidth}
    {\includegraphics[width=\linewidth]{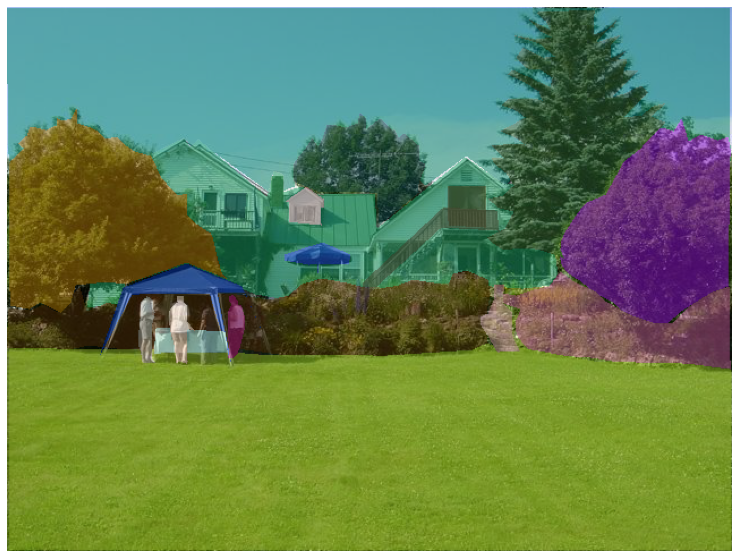}}
    \caption{GT (21)}
 \end{subfigure}
 \begin{subfigure}[b]{0.24\linewidth}
    {\includegraphics[width=\linewidth]{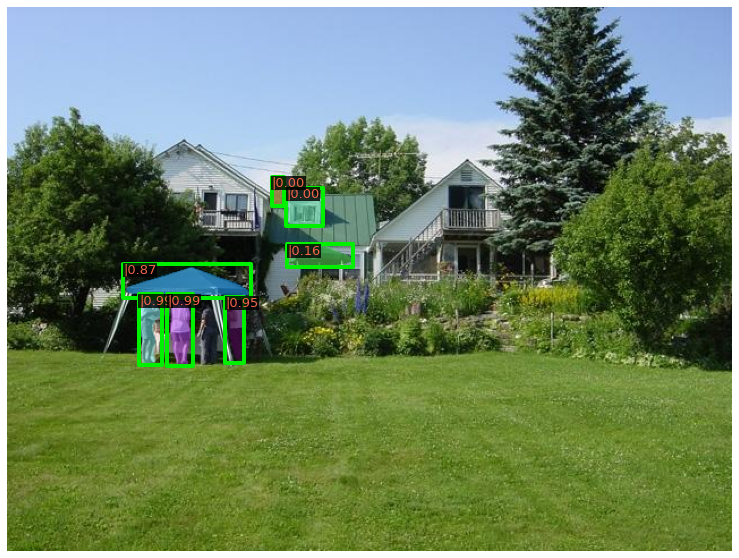}}
    \caption{Baseline (7)}
 \end{subfigure}
 \begin{subfigure}[b]{0.24\linewidth}
    {\includegraphics[width=\linewidth]{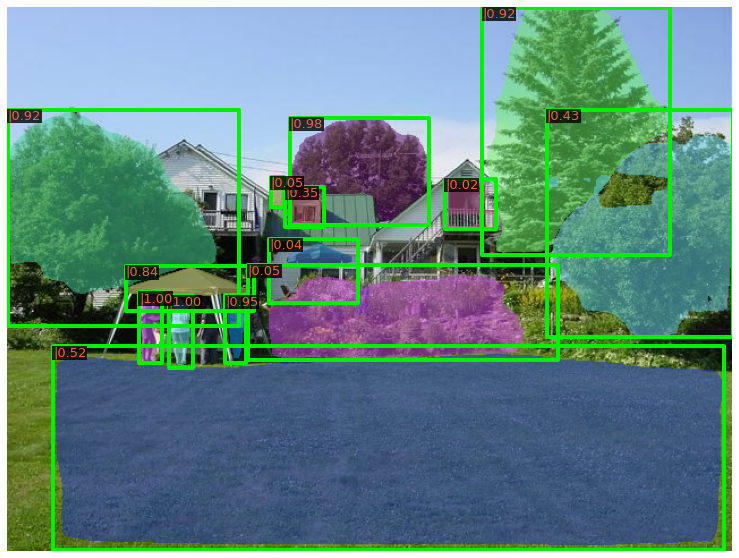}}
    \caption{\Our{} (14)}
 \end{subfigure}
 
 \begin{subfigure}[b]{0.24\linewidth}
    {\includegraphics[width=\linewidth]{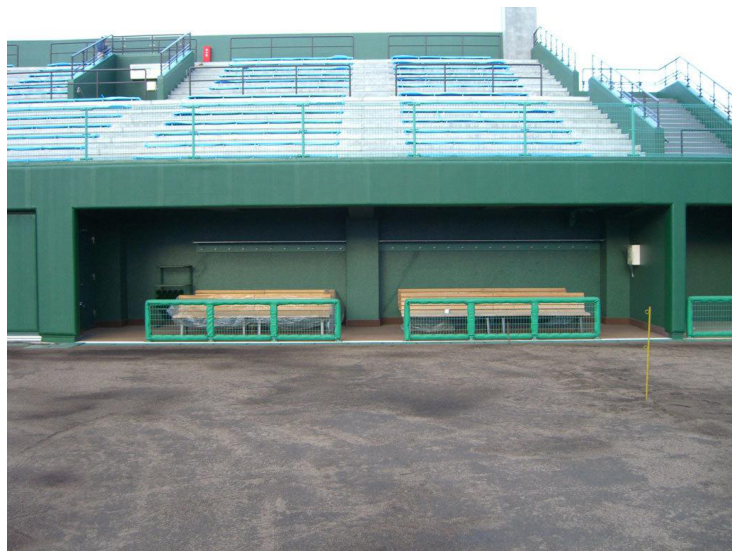}}
    \caption{Original Image}
 \end{subfigure}
 \begin{subfigure}[b]{0.24\linewidth}
    {\includegraphics[width=\linewidth]{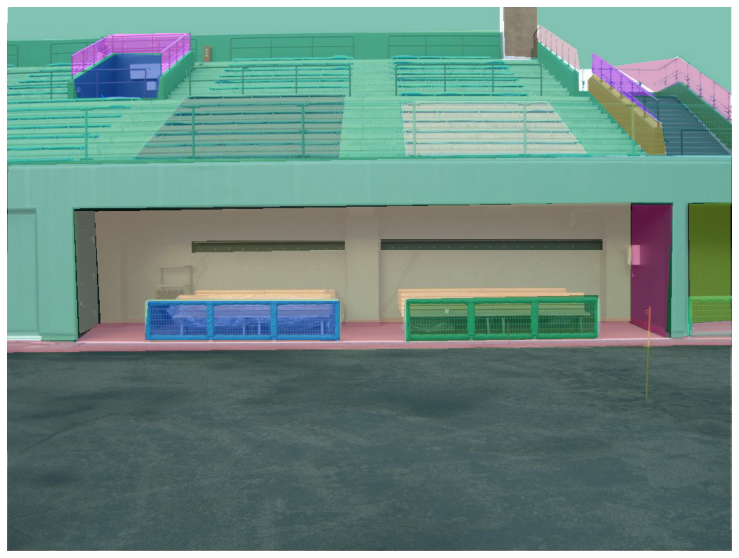}}
    \caption{GT (24)}
 \end{subfigure}
 \begin{subfigure}[b]{0.24\linewidth}
    {\includegraphics[width=\linewidth]{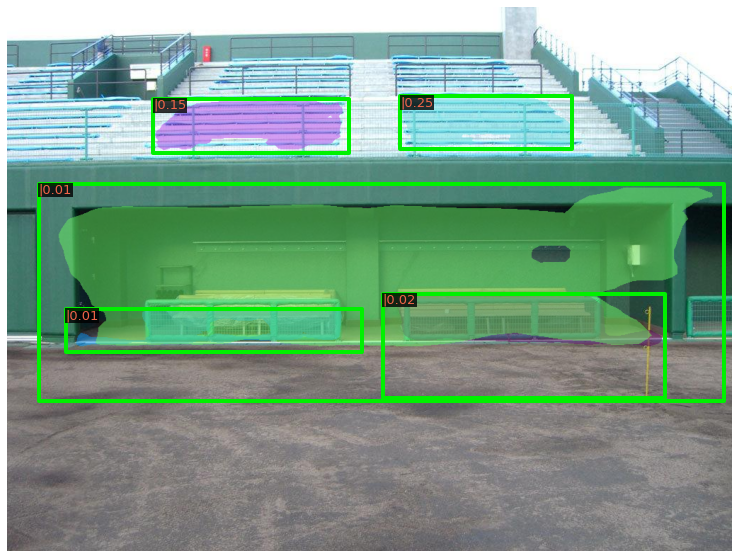}}
    \caption{Baseline (5)}
 \end{subfigure}
 \begin{subfigure}[b]{0.24\linewidth}
    {\includegraphics[width=\linewidth]{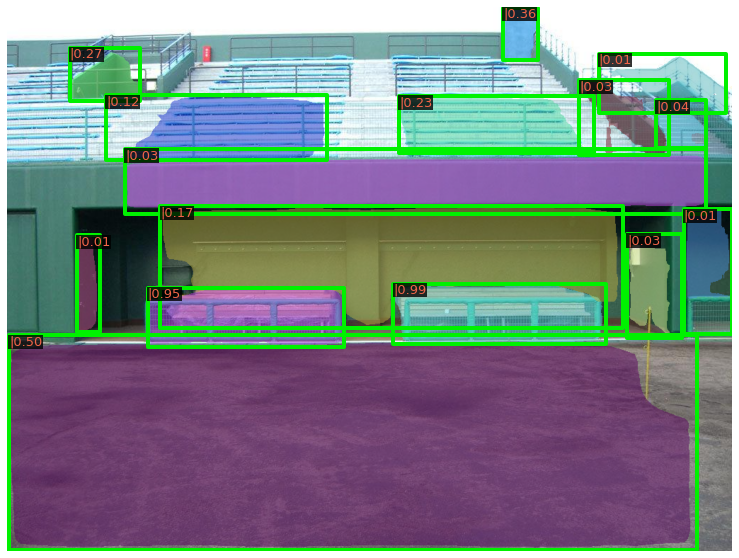}}
    \caption{\Our{} (15)}
 \end{subfigure}
 
 \begin{subfigure}[b]{0.24\linewidth}
    {\includegraphics[width=\linewidth]{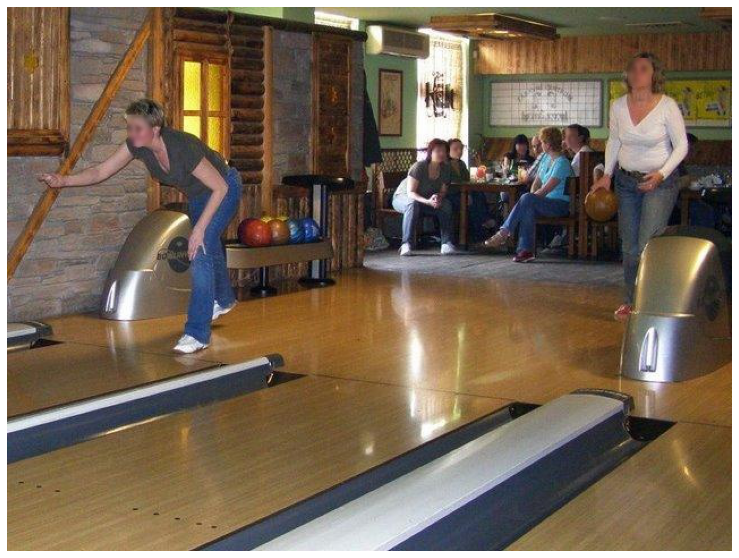}}
    \caption{Original Image}
 \end{subfigure}
 \begin{subfigure}[b]{0.24\linewidth}
    {\includegraphics[width=\linewidth]{latex/images/ade_val_vis/idx76/GT_81.png}}
    \caption{GT (81)}
 \end{subfigure}
 \begin{subfigure}[b]{0.24\linewidth}
    {\includegraphics[width=\linewidth]{latex/images/ade_val_vis/idx76/Base_mask.png}}
    \caption{Baseline (16)}
 \end{subfigure}
 \begin{subfigure}[b]{0.24\linewidth}
    {\includegraphics[width=\linewidth]{latex/images/ade_val_vis/idx76/GGN_mask.png}}
    \caption{\Our{} (30)}
 \end{subfigure}
 
 \begin{subfigure}[b]{0.24\linewidth}
    {\includegraphics[width=\linewidth]{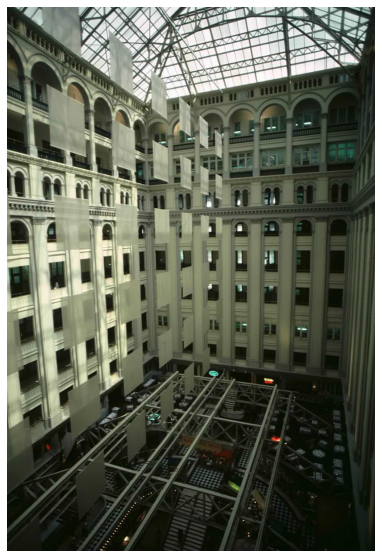}}
    \caption{Original Image}
 \end{subfigure}
 \begin{subfigure}[b]{0.24\linewidth}
    {\includegraphics[width=\linewidth]{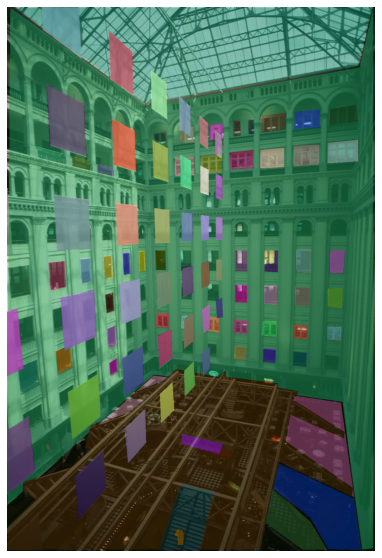}}
    \caption{GT (97)}
 \end{subfigure}
 \begin{subfigure}[b]{0.24\linewidth}
    {\includegraphics[width=\linewidth]{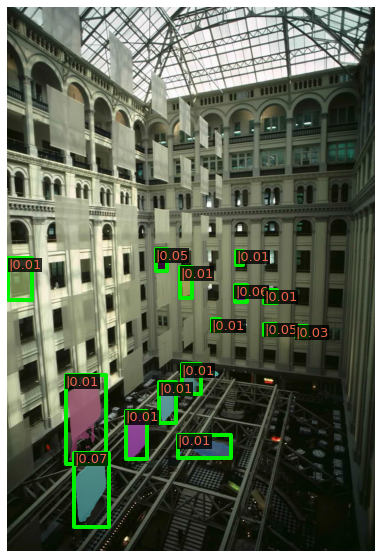}}
    \caption{Baseline (15)}
 \end{subfigure}
 \begin{subfigure}[b]{0.24\linewidth}
    {\includegraphics[width=\linewidth]{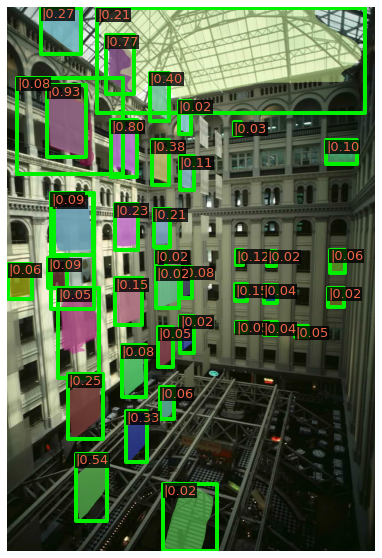}}
    \caption{\Our{} (40)}
 \end{subfigure}
 \vspace{-8pt}
 \caption{{\bf Visualization of \Our{} compared to baseline on ADE20k.} We take top-100 scoring predictions for each of the methods. \Our{} detects significantly more true positive segments compared to baseline, including novel objects and stuff. Number in bracket represents number of retrieved segments.
 }
 \label{fig:vis_ade}
 }
\end{figure*}

\begin{figure*}
\centering
{\small 
 \begin{subfigure}[b]{0.24\linewidth}
    {\includegraphics[width=\linewidth]{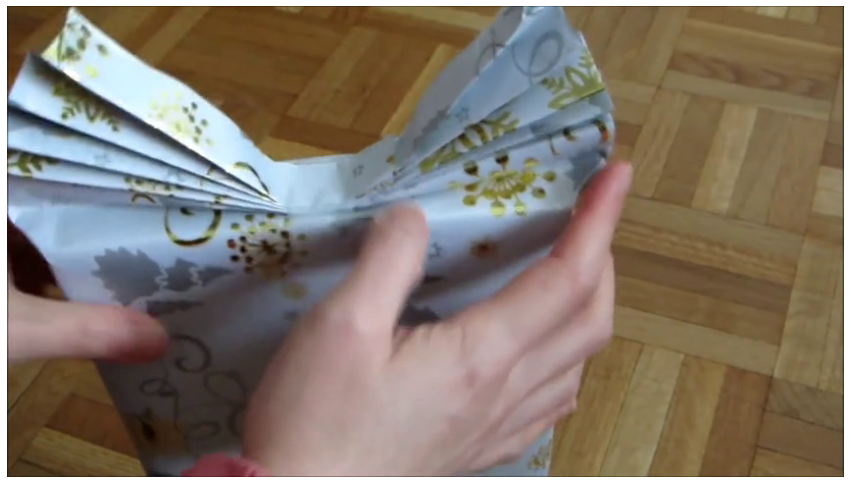}}
    \caption{Original Image}
 \end{subfigure}
 \begin{subfigure}[b]{0.24\linewidth}
    {\includegraphics[width=\linewidth]{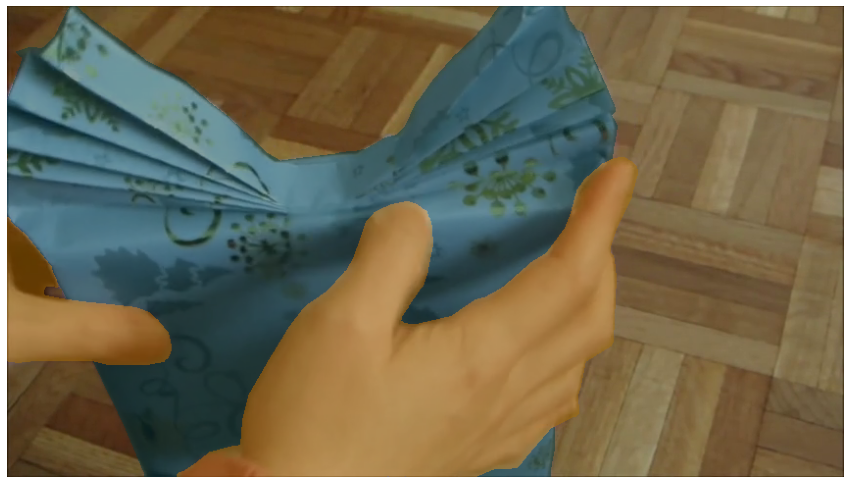}}
    \caption{GT (2)}
 \end{subfigure}
 \begin{subfigure}[b]{0.24\linewidth}
    {\includegraphics[width=\linewidth]{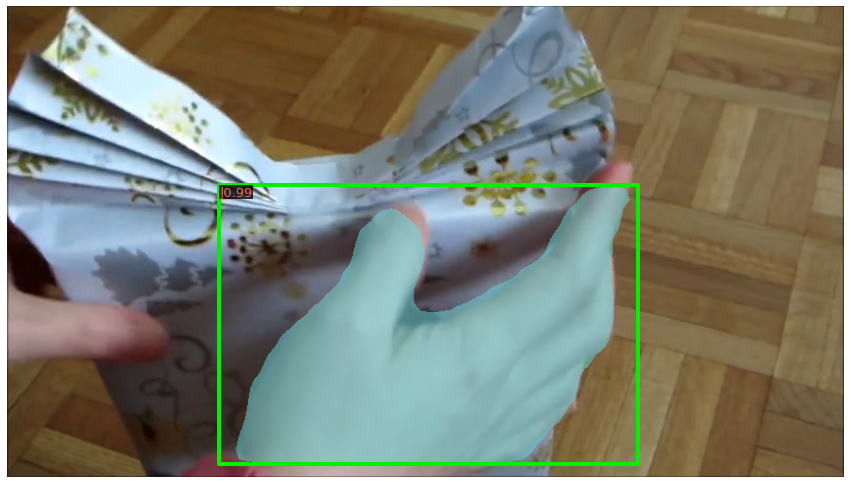}}
    \caption{Baseline (1)}
 \end{subfigure}
 \begin{subfigure}[b]{0.24\linewidth}
    {\includegraphics[width=\linewidth]{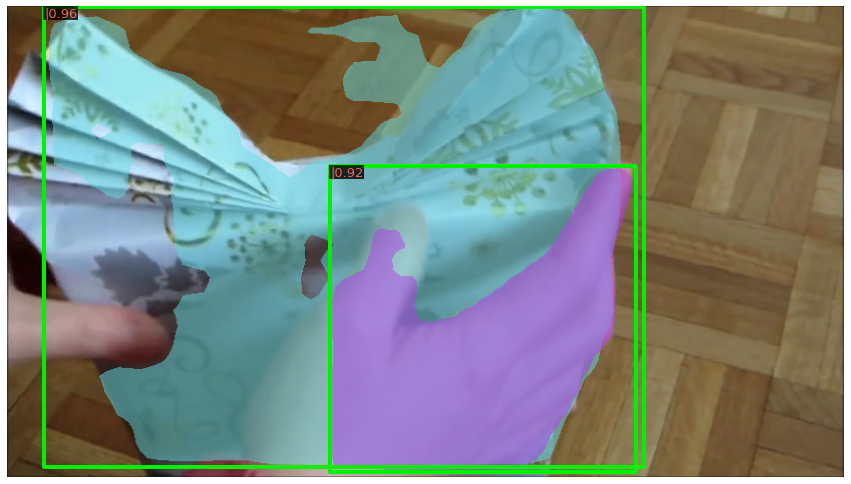}}
    \caption{\Our{} (2)}
 \end{subfigure}
 
 \begin{subfigure}[b]{0.24\linewidth}
    {\includegraphics[width=\linewidth]{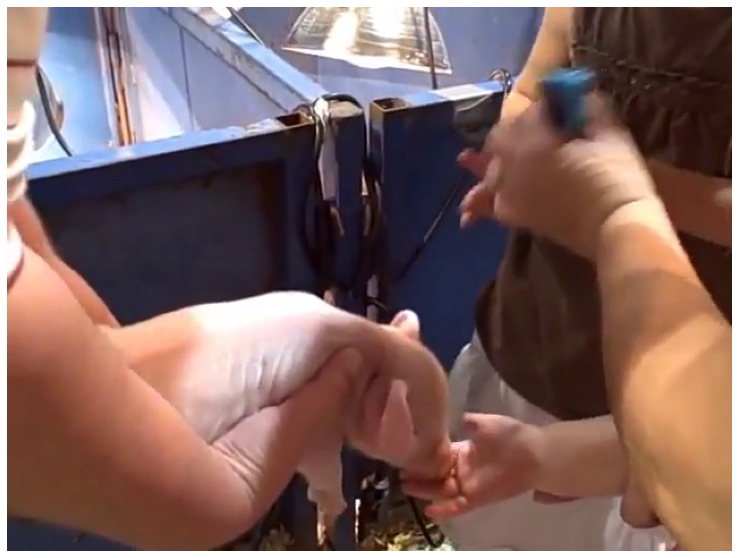}}
    \caption{Original Image}
 \end{subfigure}
 \begin{subfigure}[b]{0.24\linewidth}
    {\includegraphics[width=\linewidth]{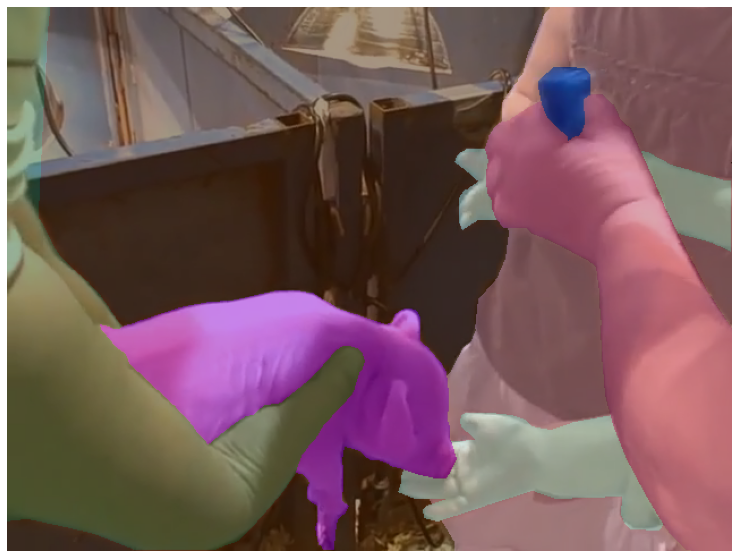}}
    \caption{GT (7)}
 \end{subfigure}
 \begin{subfigure}[b]{0.24\linewidth}
    {\includegraphics[width=\linewidth]{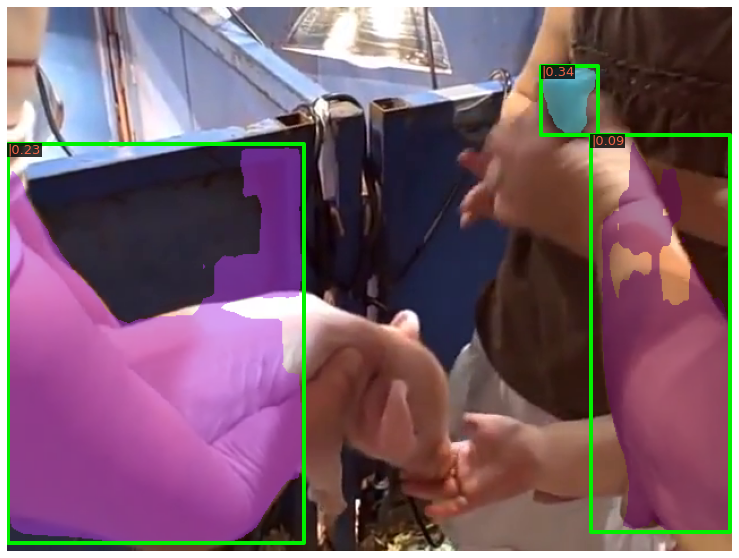}}
    \caption{Baseline (3)}
 \end{subfigure}
 \begin{subfigure}[b]{0.24\linewidth}
    {\includegraphics[width=\linewidth]{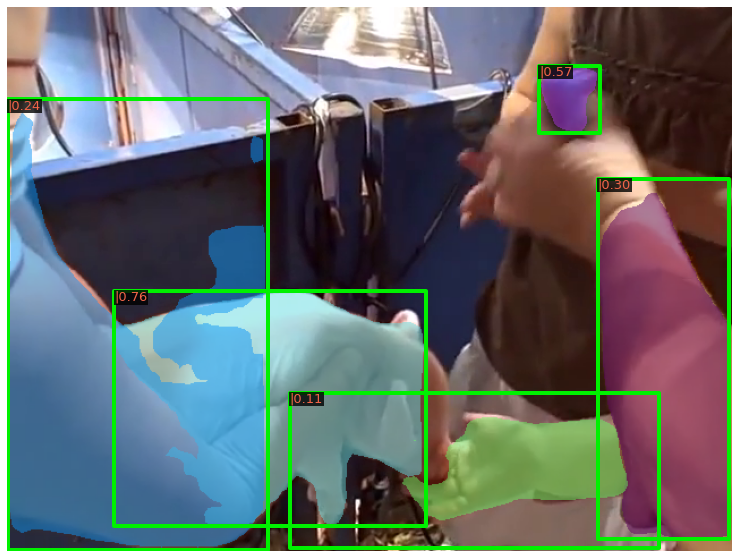}}
    \caption{\Our{} (5)}
 \end{subfigure}
 
 
 \begin{subfigure}[b]{0.24\linewidth}
    {\includegraphics[width=\linewidth]{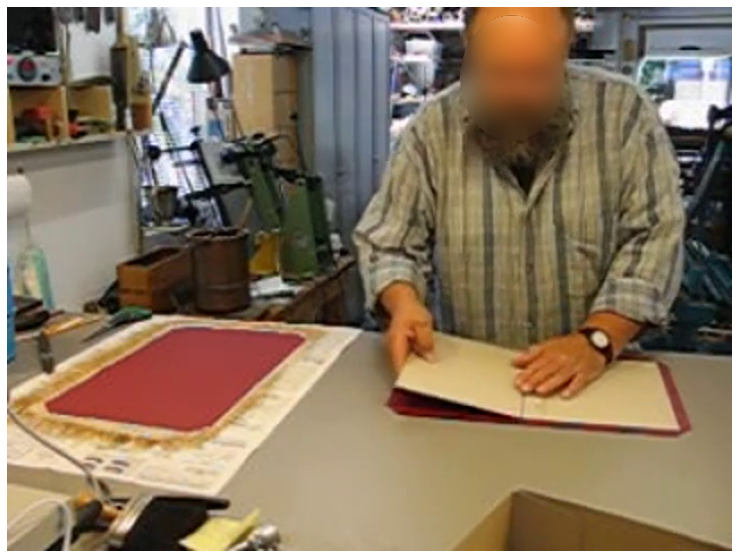}}
    \caption{Original Image}
 \end{subfigure}
 \begin{subfigure}[b]{0.24\linewidth}
    {\includegraphics[width=\linewidth]{latex/images/uvo_val_vis/idx1884/GT_13.png}}
    \caption{GT (13)}
 \end{subfigure}
 \begin{subfigure}[b]{0.24\linewidth}
    {\includegraphics[width=\linewidth]{latex/images/uvo_val_vis/idx1884/Base_mask.png}}
    \caption{Baseline (4)}
 \end{subfigure}
 \begin{subfigure}[b]{0.24\linewidth}
    {\includegraphics[width=\linewidth]{latex/images/uvo_val_vis/idx1884/GGN_mask.png}}
    \caption{\Our{} (9)}
 \end{subfigure}
 
 \begin{subfigure}[b]{0.24\linewidth}
    {\includegraphics[width=\linewidth]{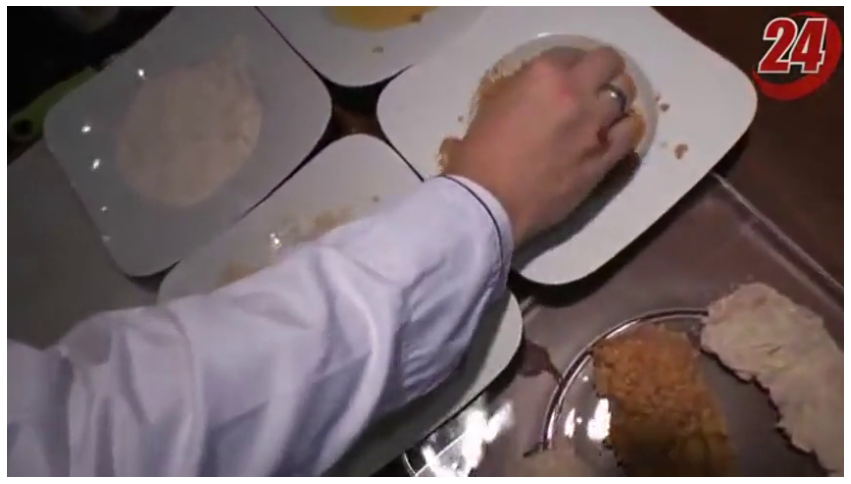}}
    \caption{Original Image}
 \end{subfigure}
 \begin{subfigure}[b]{0.24\linewidth}
    {\includegraphics[width=\linewidth]{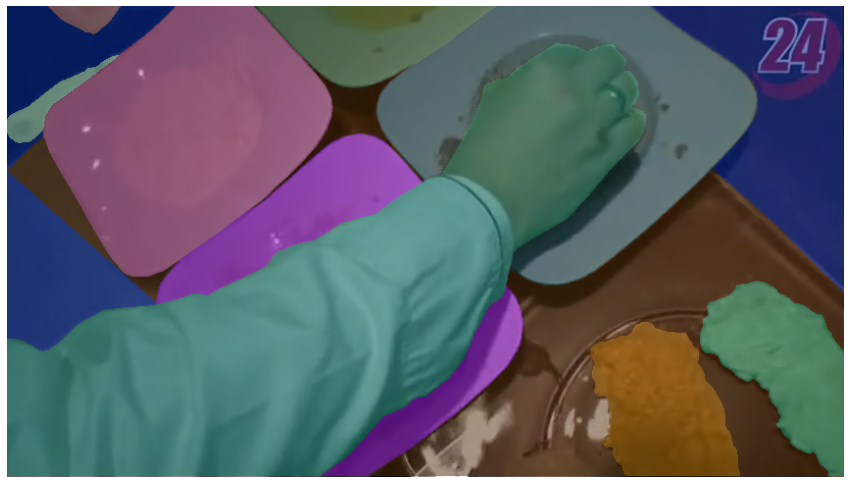}}
    \caption{GT (11)}
 \end{subfigure}
 \begin{subfigure}[b]{0.24\linewidth}
    {\includegraphics[width=\linewidth]{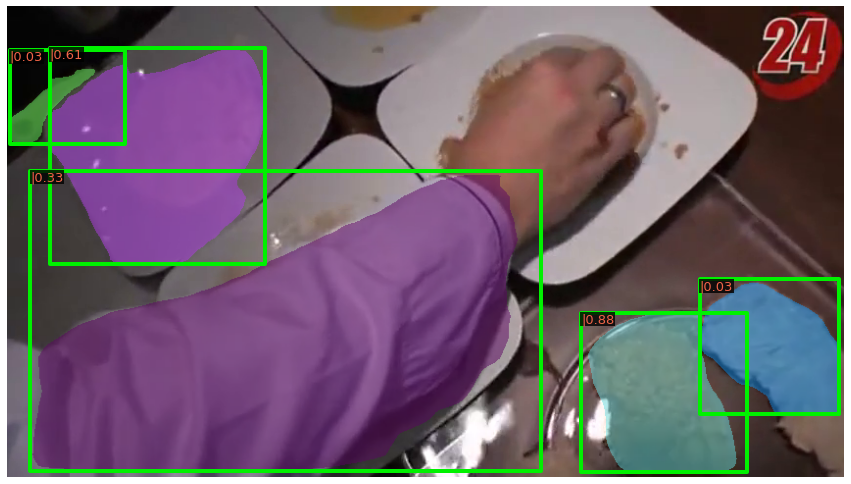}}
    \caption{Baseline (5)}
 \end{subfigure}
 \begin{subfigure}[b]{0.24\linewidth}
    {\includegraphics[width=\linewidth]{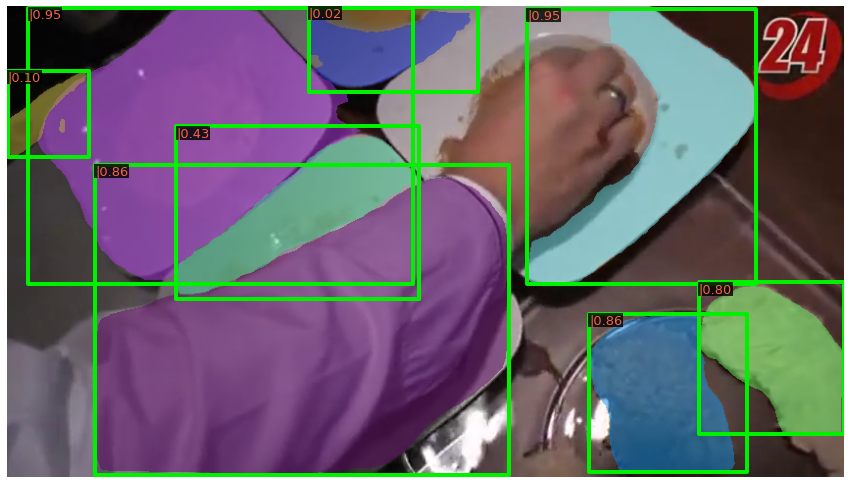}}
    \caption{\Our{} (8)}
 \end{subfigure}
 \vspace{-8pt}
 \caption{{\bf Visualization of \Our{} compared to baseline on UVO.} We take top-100 scoring predictions for each of the methods. \Our{} detects significantly more true positive segments compared to baseline, including novel objects and stuff. Number in bracket represents number of retrieved segments.
 }
 \label{fig:vis_uvo}
 }
\end{figure*}

We provide additional visualizations to compare \Our{} and baseline Mask R-CNN on ADE20k and UVO~\cite{uvo} (Figure~\ref{fig:vis_ade} and Figure~\ref{fig:vis_uvo}). Both models are trained with masks from 80 COCO categories, with \Our{} enhanced by pseudo-masks on COCO images. We show that \Our{} can recall more true positive segments than baseline, including novel objects, severely occluded objects and stuff.

\section{Does generic grouping help closed-world segmentation?}
\label{sec:two_tower}

In the previous experiments, we showed that \Our{} is useful for instance segmentation in the open-world (a.k.a class-aware instance segmentation). 
One may wonder if \Our{} is also useful for closed-world segmentation. In order to answer the question, we conduct the following proof-of-concept experiment. We adopt the standard Mask R-CNN~\cite{8237584} by replacing its RPN branch with our \Our{}. We note that our \Our{} also outputs bounding boxes and masks thus can completely replace RPN. In our experiment, \Our{} is pretrained with pseudo-GT in a class-agnostic and fixed (no fine-tuning or refinement) during class-aware training and evaluation. This means, during class-aware training and evaluation, only the recognition head is trained. We hypothesis the \Our{} can be competitive with RPN, even with a closed-world, class-aware setup. We name this modified architecture as \emph{Two-Tower} to reflex the recognition and grouping branches. The grouping branch, \Our{}, is trained on only VOC-category masks. We compare this Two-Tower architecture with Mask R-CNN which is trained end-to-end in limited data domain: using only 10\% of COCO images on all classes. Whereas Mask R-CNN is trained on grouping from all categories, the two-tower grouping module only leverages VOC masks and generated pseudo masks. Results are presented in Table~\ref{tab:two_tower}.


\begin{table}
\captionsetup{font=footnotesize}
  \centering
  {\small
  \begin{tabular}{|c|c|c|c|}
    \hline
    Training length & Method & mAP & mAR  \\
    \hline
    \multirow{2}{*}{short} & Mask R-CNN & 10.6 & 36.2 \\
     & Two-Tower & \textbf{12.7} & \textbf{36.5} \\
    \hline
    \multirow{2}{*}{normal} & Mask R-CNN & \textbf{15.4} & \textbf{40.0} \\
     & Two-Tower & 13.5 & 36.5 \\
    \hline
  \end{tabular}}
  \caption{\textbf{Proof of concept on Two-Tower model for grouping and recognition} Mask R-CNN is trained on all 80 COCO categories. \Our{}, as the grouping module, is only trained on 20 VOC classes. The recognition module does not alter the mask predictions of the grouping module, and is trained with 80 COCO categories for classification. Two-tower is competitive in both short and normal training schedules.}
  \label{tab:two_tower}
\end{table}

\section{Limitations and future directions}
\label{sec:limit}

We present \Our{} that combines bottom-up grouping and top-down training for open-world instance segmentation. The framework has shown significant gains and achieves the new state-of-the-art results on multiple benchmarks. In this section, we discuss the limitations of the approach, which also inform future directions to tackle.

\noindent \textbf{Objectness.} In \Our{}, we used WT+UCM~\cite{ucm} to group pixels into segments leveraging learned pixel pairwise affinities. However, WT+UCM has certain limitations: it has no notion of objectness, and therefore constructs pixel groups of ``part'' of an object. It is important to find novel methods to select good masks from all proposed pseudo-masks leveraging certain objectness prior, which can be learned~\cite{oln} or hand-crafted~\cite{ArbelaezPBMM14}.

\noindent \textbf{Hierarchy of groups.} When we select pseudo-GT masks generated from pairwise affinities, we ignore the natural hierarchical structure of the groups generated by UCM. It is worth understanding if enforcing grouping hierarchies can further improve the supervision signals.

\noindent \textbf{Grouping as pretext task.} Existing frameworks, such as Mask R-CNN, leverage recognition as pre-training for grouping (e.g., by pre-training on ImageNet). In this paper, we have demonstrated the value of training on unlabeled data to form grouping. A extension of this work should study how learning to group can potentially benefit recognition ability.

\section{Data augmentation for learning PA} \label{sec:PA_data_aug}

While data augmentation is well-explored in learning object proposals or masks~\cite{Zoph2020LearningDA,sohn2020detection}, it is not well-studied in the context of pairwise affinities or similar representation such as semantic edges~\cite{Liu2018SemanticED,AcunaCVPR19STEAL}. Different from bounding boxes or masks, pairwise affinities are local features and can be very sensitive to both pixel-level and spatial-level transforms. 

\begin{figure}
\centering
{\small 
 \begin{subfigure}[b]{0.49\linewidth}
    {\includegraphics[width=\linewidth]{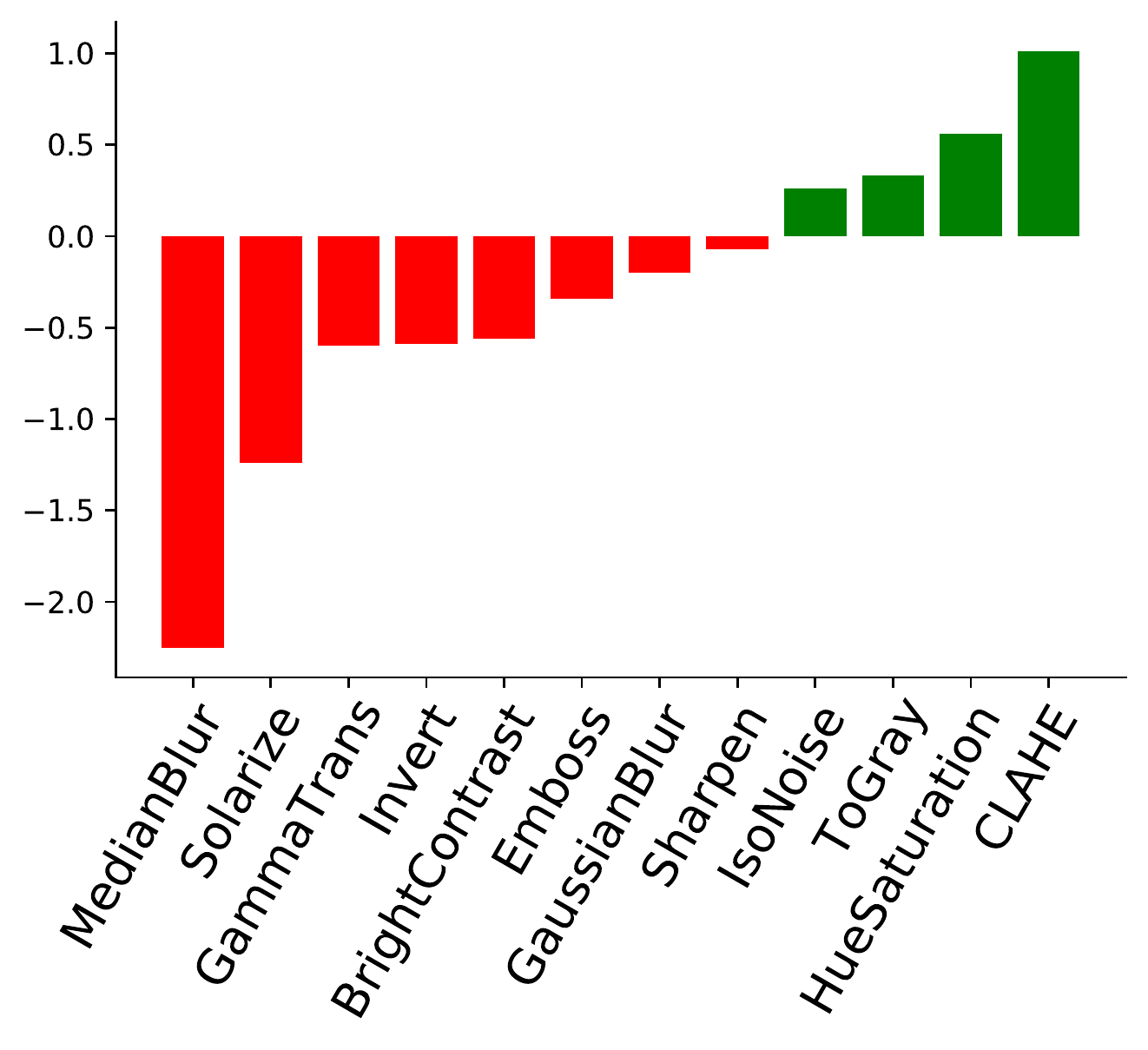}}
    \caption{}
 \end{subfigure}
 \begin{subfigure}[b]{0.49\linewidth}
    {\includegraphics[width=\linewidth]{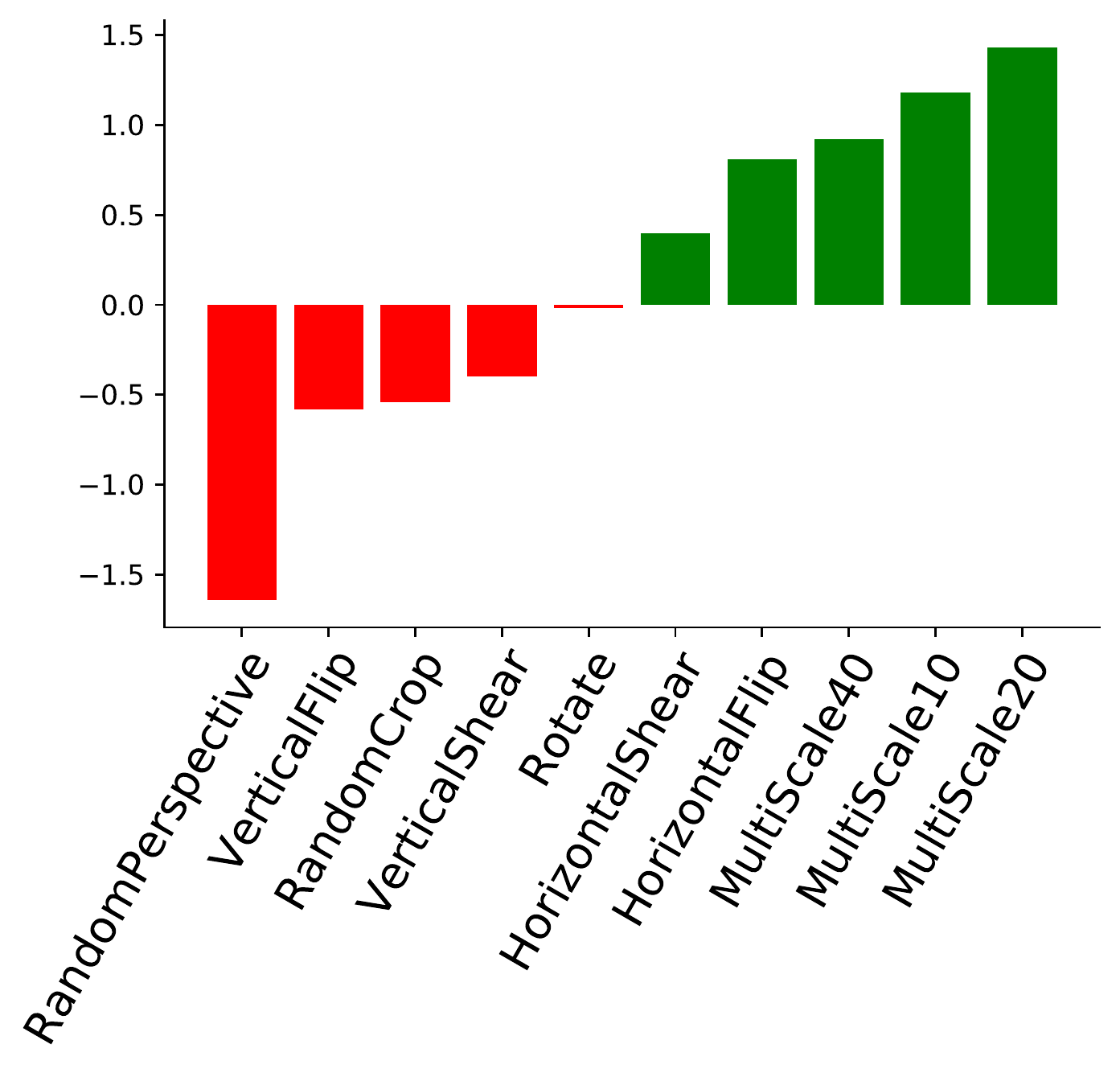}}
    \caption{}
 \end{subfigure}
 \vspace{-8pt}
 \caption{{\bf The effects of different data augmentations on learning pairwise affinities.} The performance is evaluated by UCM masks generated by the pairwise affinities trained under different augmentations. Performance is represented as gain (loss) in AR100 compared to without augmentation.  
 }
 \label{fig:data_aug}
 }
 \vspace{-8pt}
\end{figure}

\begin{figure}
\centering
{\small 
 \begin{subfigure}[b]{0.32\linewidth}
    {\includegraphics[width=\linewidth]{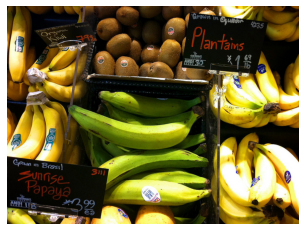}}
    \caption{original image}
 \end{subfigure}
 \begin{subfigure}[b]{0.32\linewidth}
    {\includegraphics[width=\linewidth]{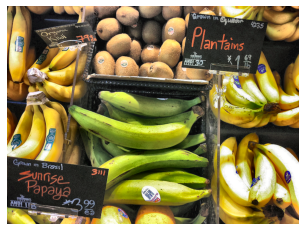}}
    \caption{CLAHE transform}
 \end{subfigure}
 \begin{subfigure}[b]{0.32\linewidth}
    {\includegraphics[width=\linewidth]{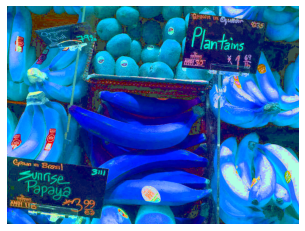}}
    \caption{HueSaturation jitter}
 \end{subfigure}
 \begin{subfigure}[b]{0.32\linewidth}
    {\includegraphics[width=\linewidth]{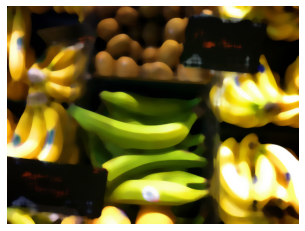}}
    \caption{Median blur}
 \end{subfigure}
 \begin{subfigure}[b]{0.32\linewidth}
    {\includegraphics[width=\linewidth]{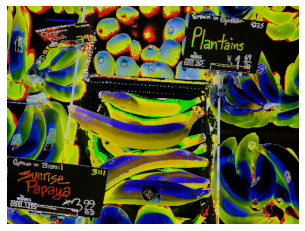}}
    \caption{Solarize}
 \end{subfigure}
 \begin{subfigure}[b]{0.32\linewidth}
    {\includegraphics[width=\linewidth]{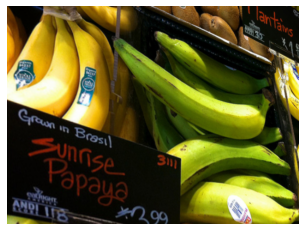}}
    \caption{Random perspective}
 \end{subfigure}
 \vspace{-8pt}
 \caption{{\bf Visualization of data augmentation.} Top row includes original image and two strong pixel-level augmentation. Bottom row contains three augmentation types that hurt the performance. 
 }
 \label{fig:vis_aug}
 }
 \vspace{-8pt}
\end{figure}

Many data augmentation has a positive effect on pairwise affinities: multi-scaling is the strongest augmentation among all. Besides, CLAHE~\cite{PIZER1987355} and Hue-Saturation value jittering provide strong pixel-level augmentation. Not all augmentation helps to learn pairwise affinities: among 20 types of augmentation experimented, more than half hurts the performance of pairwise affinities (Fig.~\ref{fig:data_aug}). For instance, different from the findings in contrastive learning~\cite{He2020MomentumCF,chen2020simple}, all kinds of blurring hurts the performance of pairwise affinities. Pairwise affinities predict local relationship, which becomes uncertain with blurred images. In addition, orientation matters. While horizontal flipping and shearing contribute positively to learning pairwise affinities, vertical operators of the same kinds hurt the performance. We visualize a few augmentation in Figure~\ref{fig:vis_aug}.

\end{document}